\documentclass[10pt,twocolumn,letterpaper]{article}

\usepackage[pagenumbers]{iccv} 

\newcommand{\textitgray}[1]{\textit{\textcolor{gray}{#1}}}
\newcommand{\redtext}[1]{\textcolor[RGB]{205,38,38}{#1}}
\newcommand{\greentext}[1]{\textcolor[RGB]{34, 139, 34}{#1}}

\usepackage{amsmath}
\usepackage{amssymb}
\usepackage{booktabs}
\usepackage{wrapfig}

\usepackage{pifont}
\usepackage{multirow}
\usepackage{multicol}
\usepackage{booktabs} 
\usepackage{rotating} 
\usepackage{makecell} 
\usepackage{tikz}

\usepackage[sectionbib]{chapterbib}

\definecolor{iccvblue}{rgb}{0.21,0.49,0.74}
\usepackage[pagebackref,breaklinks,colorlinks,allcolors=iccvblue]{hyperref}

\def\paperID{7355} 
\def\confName{ICCV}
\def\confYear{2025}

\title{GigaSLAM: Large-Scale Monocular SLAM with Hierarchical Gaussian Splats}

\author{
Kai Deng\textsuperscript{1}, Yigong Zhang\textsuperscript{1}, Jian Yang\textsuperscript{1, 2}, Jin Xie\textsuperscript{1, 2}\\
\textsuperscript{1}Nankai University, Tianjin, China\\
\textsuperscript{2}Nanjing University Suzhou Campus, Suzhou, China\\
{\tt\small dengkai@mail.nankai.edu.cn, zyg025@nankai.edu.cn, csjyang@nankai.edu.cn, csjxie@nju.edu.cn}
}

\begin{document}

\twocolumn[{%
\renewcommand\twocolumn[1][]{#1}%
\maketitle

\begin{center}
  \centering
   \includegraphics[width=\linewidth]{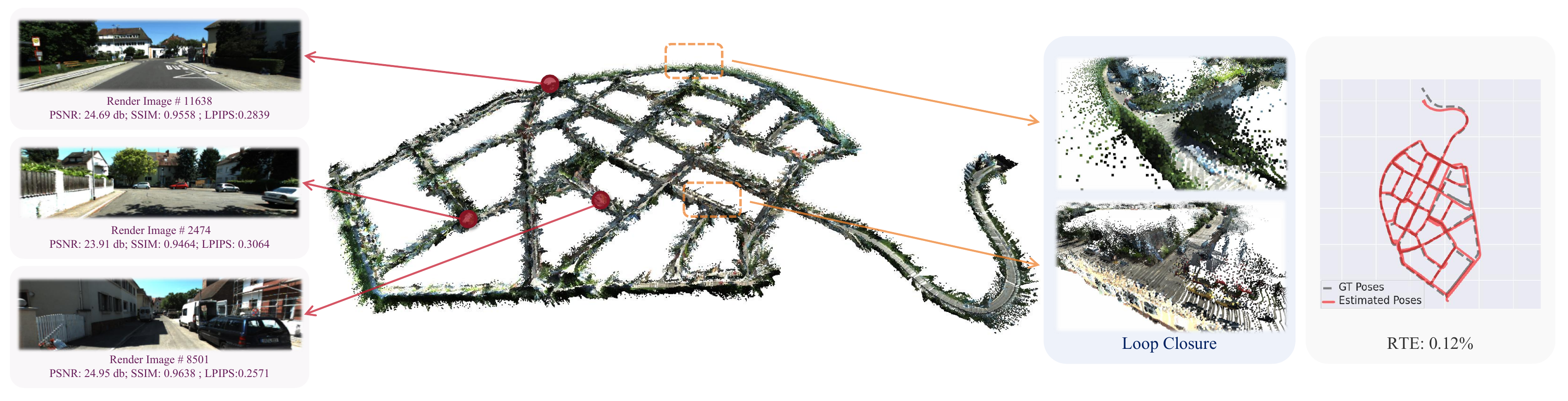}

   \captionof{figure}{
      {\bf GigaSLAM} achieves robust pose estimation and mapping accuracy across {unbounded, multi-kilometer-scale outdoor sequences} while preserving high-fidelity scene rendering quality, highlighting the effectiveness of our approach for long-range, real-world scenarios.
   }
   \label{fig:overview}
\end{center}
}]

\begin{abstract}

Tracking and mapping in large-scale, unbounded outdoor environments using \textbf{only monocular RGB input} presents substantial challenges for existing SLAM systems. Traditional Neural Radiance Fields (NeRF) and 3D Gaussian Splatting (3DGS) SLAM methods are typically limited to small, bounded indoor settings. To overcome these challenges, we introduce \textbf{GigaSLAM}\footnote{\scriptsize https://github.com/DengKaiCQ/GigaSLAM}, the \textbf{first RGB} NeRF / 3DGS-based SLAM framework for \textbf{kilometer-scale outdoor environments}, as demonstrated on the KITTI, KITTI 360, 4 Seasons and A2D2 datasets. Our approach employs a hierarchical sparse voxel map representation, where Gaussians are decoded by neural networks at multiple levels of detail. This design enables efficient, scalable mapping and high-fidelity viewpoint rendering across expansive, unbounded scenes. For front-end tracking, GigaSLAM utilizes a metric depth model combined with epipolar geometry and PnP algorithms to accurately estimate poses, while incorporating a Bag-of-Words-based loop closure mechanism to maintain robust alignment over long trajectories. Consequently, GigaSLAM delivers high-precision tracking and visually faithful rendering on urban outdoor benchmarks, establishing a robust SLAM solution for large-scale, long-term scenarios, and significantly extending the applicability of Gaussian Splatting SLAM systems to unbounded outdoor environments.

\end{abstract}    
\section{Introduction}
\label{sec:intro}

Simultaneous localization and mapping (SLAM) from a single monocular video is a longstanding challenge in computer vision \cite{slampast}. Conventional SLAM approaches \cite{orbslam,orbslam2,orbslam3, LSD-SLAM} seek accurate camera pose tracking and high-quality map geometry. Recent advances in neural radiance fields (NeRF) \cite{mildenhall2021nerf} and 3D gaussian splatting (3DGS) \cite{kerbl20233gs} have inspired the SLAM community \cite{tosi2024nerfs} to enhance map capabilities—enabling rich appearance encoding and realistic rendering from free viewpoints. Novel SLAM techniques using NeRFs and 3DGS have shown great promise for online mapping and rendering \cite{zhu2022nice,zhu2023nicer,zhang2023go,keetha2024splatam,matsuki2024gaussian}. These capabilities open new possibilities for applications in AR/VR, autonomous driving, and drone navigation by allowing users and agents to render from different perspectives. Unlike traditional 3D GS or NeRF methods that rely on a pre-reconstructed static scene from SfM\cite{colmap}, SLAM-based approaches must dynamically build and update the 3D scene online. This online nature introduces unique challenges, especially when handling loop closures and global scene corrections.

Despite these advancements, one key challenge remains for such systems: current NeRF and 3DGS-based SLAM frameworks are often limited in scene scale, with most relying on RGB-D depth priors for accurate mapping and tracking in larger scenes \cite{zhu2022nice,keetha2024splatam,matsuki2024gaussian}. This is due to two main reasons: the limitations of scene representations and challenges in global alignment. Implicit methods, such as NeRFs \cite{zhu2022nice,zhu2023nicer,zhang2023go}, have limited representational capacity and are often confined to bounded regions. Their reliance on manually pre-defined scene bounding boxes becomes impractical in expansive outdoor environments with dynamic scales and undefined boundaries. Furthermore, the prevalent dense volumetric grid representations exhibit cubic space complexity $O(n^3)$, which incurs prohibitive memory and computational costs when scaling to outdoor scenes spanning thousands of cubic meters. Explicit methods, like Gaussian splatting \cite{keetha2024splatam,matsuki2024gaussian}, are not memory-efficient, as their size scales with scene growth, impacting computational and memory efficiency. Meanwhile, concurrent work like OpenGS-SLAM \cite{opengs-slam} integrates 3R modules \cite{wang2024dust3r} with 3D GS, achieving capability on short hundred-meter-scale outdoor sequences (Waymo\cite{waymo}). However, its Transformer-based 3R design incurs $O(n^2)$ memory costs due to self-attention, limiting scalability to kilometer-scale trajectories. Current 3R-based forward 3D reconstruction frameworks remain limited in scalability, as no prior method has demonstrated robust performance on sequences exceeding kilometer-level spatial extents. The second challenge lies in global alignment. In large-scale scenes, loop closure is crucial, as it effectively reduces global drift. However, most NeRF and 3DGS SLAM methods rely on incremental gradient-based registration, which is not well-suited for global alignment under such scenarios. As a result, existing NeRF and 3DGS SLAM methods are restricted to small-scale, bounded indoor environments or short-term outdoor scenes and often depend on RGB-D data for robust scene alignment.

To address these challenges, we present GigaSLAM, a Gaussian Splatting-based SLAM framework designed to scale to large, outdoor, long-term, unbounded environments. The core of GigaSLAM's technical contribution is a novel hierarchical sparse voxel representation designed for large-scale rendering-capable SLAM, with each level encoding Gaussians at varying levels of detail. This map representation has two key advantages for scaling: 1) it is boundless, dynamically expanding as the camera moves; 2) it enables content-aware, efficient rendering at large scale through Level-of-Details (LoD) representation, reducing the need to load hundreds of millions of Gaussians for a single frame. We further enhance map geometry and camera pose tracking accuracy using a data-driven monocular metric depth module \cite{piccinelli2024unidepth}. Finally, we integrate Bag-of-Words loop closure detection \cite{galvez2012bags} and design a comprehensive post-closure Splats map update to maintain accuracy in rendering. GigaSLAM has been shown to scale up to tens of kilometers of travel distance in urban driving scenarios \cite{geiger2012kitti,liao2022kitti, wenzel20214seasons, geyer2020a2d2}. It achieves robust and accurate tracking with the ability to render from any viewpoint, all on a single GPU. To our knowledge, no existing framework offers this capability.

We validate our methods on large-scale outdoor scenes from the KITTI, KITTI 360, 4 Seasons and A2D2 dataset \cite{geiger2012kitti,liao2022kitti, wenzel20214seasons, geyer2020a2d2}. 	Our results indicate that this approach is significantly more accurate and robust, outperforming other monocular SLAM methods in average tracking performance on long-term outdoor datasets comparing to current 3D GS SLAM methods \cite{matsuki2024gaussian, sandstrom2024splat} which are tailored for indoor scenes with monocular RGB images. Therefore, existing NeRF and 3D Gaussian Splatting-based SLAM frameworks struggle to handle these large-scale, unbounded scenarios effectively. Our contributions are as follows: 1) GigaSLAM, a novel Gaussian Splats-based SLAM framework for large-scale, unbounded environments; 2) a hierarchical map representation for dynamic growth and level-of-detail rendering in large-scale SLAM; 3) an efficient loop closure procedure applicable to Gaussian splats map representations.
\section{Related Work}
\label{sec:relatedwork}

\paragraph{Traditional SLAM}
Modern SLAM systems are typically formulated as joint optimization problems \cite{slampast}, where the goal is to estimate a robot’s pose (position and orientation) from video input. A representative system is ORB-SLAM \cite{orbslam}, along with its extended versions ORB-SLAM2 \cite{orbslam2} and ORB-SLAM3 \cite{orbslam3}, which use feature points and keyframes for monocular SLAM. In more recent studies, researchers have incorporated deep learning models to solve specific sub-modules within SLAM, which are then integrated into traditional optimization-based frameworks. For example, the authors of $\nabla$ SLAM \cite{gradslam} propose using automatic differentiation to model SLAM as a differentiable computation graph. Similarly, CodeSLAM \cite{codeslam} introduces an autoencoder-based representation for dense monocular SLAM by learning compact geometric descriptors from images. Several works \cite{banet, Pixel-perfect} have also embedded Bundle Adjustment into end-to-end differentiable networks. DROID-SLAM \cite{droidslam} exemplifies this trend by integrating Dense Bundle Adjustment directly into an optical flow estimation pipeline.

\paragraph{NeRF-based SLAM}
The introduction of Neural Radiance Fields (NeRF) \cite{mildenhall2021nerf} has inspired a range of SLAM systems that leverage neural rendering for mapping and localization. iMAP \cite{sucar2021imap} is one of the first works to explore NeRF-based scene reconstruction for SLAM. Building on this, NICE-SLAM \cite{zhu2022nice} adopts a grid-based hierarchical representation to improve efficiency in indoor mapping. NICER-SLAM \cite{zhu2023nicer} further introduces geometric and optical flow constraints, along with a warping loss, to improve consistency while supporting monocular input. The authors of Point-SLAM \cite{sandstrom2023point} employ point-based neural fields for explicit scene representation, though their method requires RGB-D input and does not integrate with traditional SLAM architectures. GO-SLAM \cite{zhang2023go} extends NICE-SLAM by incorporating loop closure detection to enhance global reconstruction consistency. However, since NeRF-based systems generally require pre-defined scene boundaries, they struggle to scale to unbounded outdoor environments, limiting their applicability in such scenarios.

\paragraph{Gaussian Splatting-based SLAM}
The emergence of 3D Gaussian Splatting (3DGS) \cite{kerbl20233gs} has motivated researchers to explore its potential for SLAM. The earliest SLAM adaptations of 3DGS include SplaTAM \cite{keetha2024splatam} and MonoGS \cite{matsuki2024gaussian}. In SplaTAM, the authors propose an online SLAM framework that employs Gaussian primitives and differentiable contour-guided optimization for both tracking and mapping. MonoGS takes advantage of the explicit and compact nature of Gaussians, introducing geometric validation and regularization to address ambiguities in dense reconstruction; the method supports both RGB and RGB-D inputs, though RGB-only input leads to reduced tracking and mapping quality. GS-SLAM \cite{yan2024gs} introduces an adaptive extension strategy for efficient map updates and reconstruction of novel regions. RGBD GS-ICP SLAM \cite{ha2024rgbd} incorporates Generalized ICP with 3DGS to improve localization precision. Splat-SLAM \cite{sandstrom2024splat}, based on the tracking framework of DROID-SLAM \cite{droidslam}, achieves state-of-the-art accuracy in indoor scenes using only RGB input, but its performance in outdoor environments remains unverified. VPGS-SLAM \cite{deng2025vpgs} is an independent work shortly after our study. It maintains LiDAR sensors as input, thereby incurring higher hardware costs typical of such LiDAR-based systems. Overall, current 3DGS-based SLAM systems tend to rely on RGB-D or LiDAR data and are mostly evaluated on indoor scenes, with limited validation on large-scale or long-term outdoor sequences.

\begin{figure*}[htbp]
  \centering
   \includegraphics[width=\linewidth]{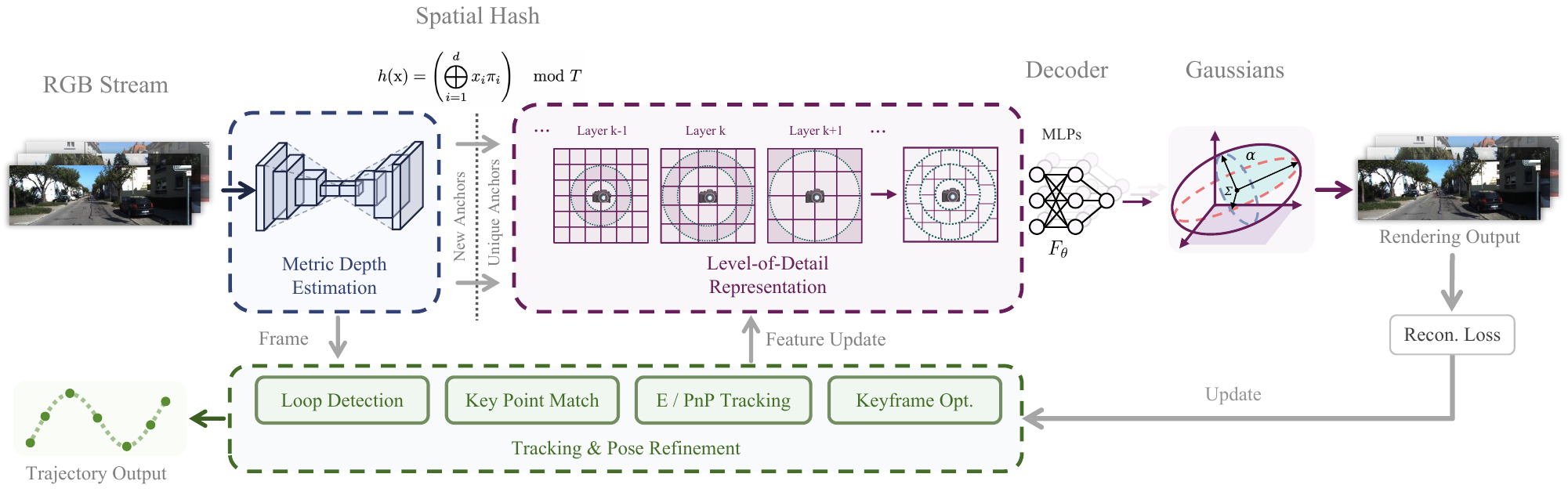}

   \caption{{\bf Overview of our algorithm.} GigaSLAM processes monocular RGB input to map large-scale outdoor environments using a hierarchical sparse voxel structure. By this structure we solve the challenging problem at long distances outdoor scenarios.}
   \label{fig:overview}
\end{figure*}

\section{Method}

GigaSLAM maps large-scale outdoor environments from monocular RGB input using a hierarchical sparse voxel structure. An overview is shown in Figure \ref{fig:overview}.

\subsection{Preliminaries}
\label{sec:splat}

\paragraph{Gaussian Splatting \cite{kerbl20233gs}}

A Gaussian primitive $\mathbf{G} = (\mathbf{c}, \mathbf{s}, \alpha, \boldsymbol{\mu}, \mathbf{\Sigma})$ is defined by its color $\mathbf{c} \in \mathbb{R}^3$, scale $\mathbf{s} \in \mathbb{R}^3$, opacity $\alpha \in \mathbb{R}$, mean vector $\boldsymbol{\mu} \in \mathbb{R}^3$, and diagonal covariance matrix $\mathbf{\Sigma}$, which represents the ellipsoidal shape and position of the Gaussian in 3D. Each Gaussian primitive $\mathbf{G}$ is derived from a sparse voxel $\mathbf{V}_{t, i}$ via a neural network decoder, represented as a multi-layer perceptron $F_\theta(\cdot)$ with learnable parameters $\theta$. At each time $t$, the map $\mathcal{M}_t = (\mathcal{V}_t, \mathcal{G}_t)$ consists of sparse voxels $\mathbf{V}_{t, i}$ and Gaussian splats $\mathcal{G}_t$, with each $\mathbf{V}_{t, i}$ denoting the $i$-th voxel at time $t$.

For rendering, the primitives are sorted by their distance to the camera, and alpha compositing is applied. Each 3D Gaussian $\mathcal{N}(\boldsymbol{\mu}, \mathbf{\Sigma})$ projects onto the image plane as a 2D Gaussian $\mathcal{N}(\boldsymbol{\mu}_I, \boldsymbol{\Sigma}_I)$:
\begin{equation}
    \boldsymbol{\mu}_I = -\pi(\boldsymbol{\xi} \cdot \boldsymbol{\mu}), \quad \Sigma_I = \mathbf{J} \mathbf{R} \mathbf{\Sigma} \mathbf{R}^T \mathbf{J}^T,
\end{equation}
where $\pi(\cdot)$ is the projection function, $\boldsymbol{\xi} \in \text{SE}(3)$ is the camera pose, $\mathbf{J}$ is the Jacobian, and $\mathbf{R}$ is the camera rotation matrix. This setup ensures end-to-end differentiability of the 3D Gaussian splatting.

The pixel color $C_\mathbf{p}$ at a pixel position $\mathbf{p} = (u, v)$ and the depth $D_\mathbf{p}^\mathrm{GS}$ at $\mathbf{p}$ are computed as:
\begingroup
\small
\begin{equation}
    C_\mathbf{p} = \sum_{i=1}^{|\mathcal{G}_t|} c_i \alpha_i \prod_{j=1}^{i-1} (1 - \alpha_j), \,
    D_\mathbf{p}^\mathrm{GS} = \sum_{i=1}^{|\mathcal{G}_t|} z_i \alpha_i \prod_{j=1}^{i-1} (1 - \alpha_j),
\end{equation}
\endgroup
where $z_i$ is the camera-to-mean distance for the $i$-th Gaussian.

\paragraph{Voxelized Gaussian Representation}

Our work leverages Scaffold-GS \cite{lu2024scaffold} for a voxelized 3D representation. This representation offers fast rendering, quick convergence, and improved memory efficiency by encoding 3DGS data into feature vectors. Additionally, it allows multiple points mapped to the same voxel to be merged into a single voxel representation, reducing memory usage by minimizing redundancy. The scene is divided into sparse voxels. The center position of each voxel is referred to as the anchor position $ \mathbf{V} \in \mathbb{R}^{N \times 3} $, each containing a context feature vector $ \hat{f_v} \in \mathbb{R}^{32} $, a scaling factor $ \mathbf{l}_v \in \mathbb{R}^3 $, and $ k $ offsets $ \mathbf{O}_v \in \mathbb{R}^{k \times 3} $. Each voxel is decoded into $ k $ Gaussians via shared MLPs. Using MLPs $ F_\alpha, F_{\text{color}}, F_{\text{quan}}, F_{\text{scaling}} $, the distance $ \delta = \Vert \mathbf{x}_v - \mathbf{x}_c \Vert_2 $ from the voxel to the camera, and the viewing direction $ \vec{\mathbf{d}}_{vc} = (\mathbf{x}_v - \mathbf{x}_c)/\delta $, the corresponding factors of Gaussian primitives are generated by,
\begin{equation}
    \begin{aligned}
\{\alpha_i\}^{k}_{i=1} &= F_\alpha(\hat{f}_v, \delta,\vec{\mathbf{d}}_{vc}),
\end{aligned}
\end{equation}
$\mathbf{c}_i$, $\mathbf{\Sigma}_i$ and $\mathbf{s}_i$ can be got in similar way and $\boldsymbol{\mu}_i$ is calculated:
\begin{equation}
    \begin{aligned}
\{\boldsymbol{\mu}_i\}^{k}_{i=1} &= \mathbf{x}_v + \{\mathbf{O}_{v,i}\}^{k}_{i=1} \cdot l_v.
\end{aligned}
\end{equation}

Once decoded by MLPs, the Gaussians are rendered through the Splat operation as described in the previous part.

\subsection{Mapping}
\label{sec:mapping}

In large-scale SLAM, the key challenge is choosing a scalable, expressive, and flexible map representation for outdoor environments.   Unlike indoor scenes, the open nature of outdoor environments makes implicit representations like NeRF impractical, as they struggle with infinite extents and dynamic depth ranges. For 3D GS, the large number of distant Gaussian primitives can reduce rendering efficiency, as they contribute little to the output but significantly increase computational load.

\paragraph{Hierarchical Representation}

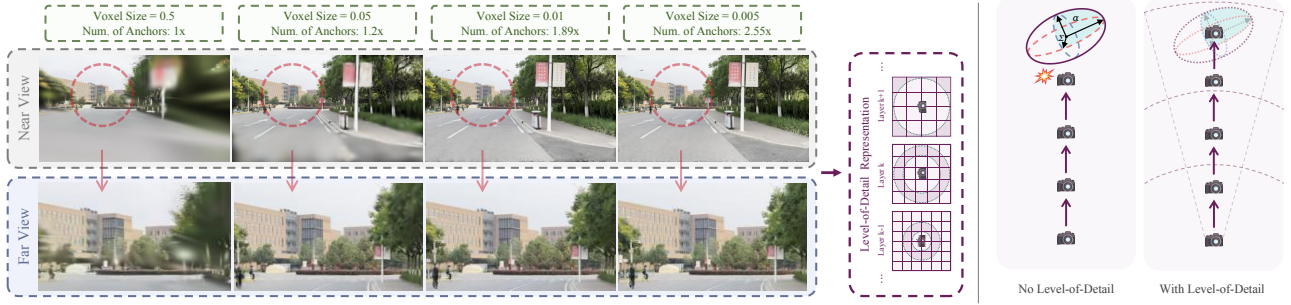
\begin{figure*}[ht]
    \centering

    \begin{subfigure}{0.74 \textwidth}
        \centering 
        \includegraphics[width=\linewidth]{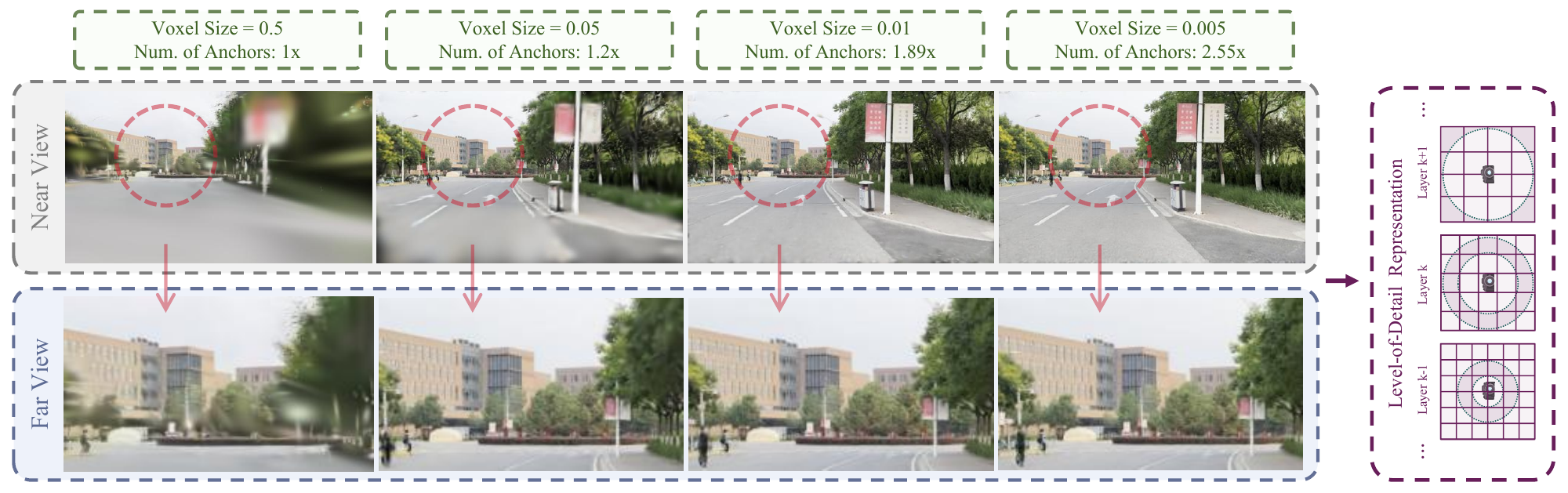}
    \end{subfigure}
    \begin{tikzpicture}
        \draw[gray] (0,0) -- (0, -4); 
    \end{tikzpicture}
    \begin{subfigure}{0.24 \textwidth}
        \centering
        \includegraphics[width=\linewidth]{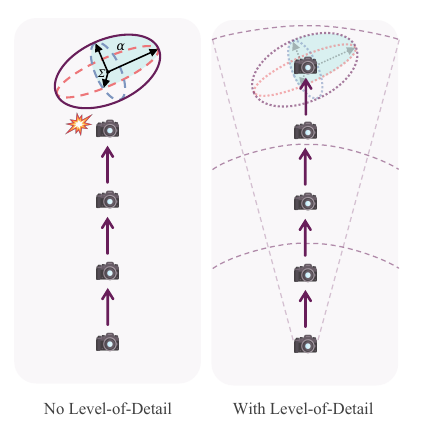}
    \end{subfigure}

    \caption{{\bf Efficiency and effectiveness of LoD Map Representation.} (Left) Voxel refinement improves reconstruction for nearby objects, with diminishing gains for distant scenes. A hierarchical 3DGS approach balances coarser and finer voxel representations. (Right) ``Collision" issue where Gaussians from distant views overlap with those in subsequent frames in a long sequence.
    }
    \label{fig:lod_large}
\end{figure*}

Using a voxel-based representation we establish a hierarchical structure by adjusting voxel size. As shown in Figure \ref{fig:lod_large} (left), increasing voxel detail requires more computational resources but provides limited improvement for distant elements like buildings or the sky. Therefore, a 3D GS representation with finer voxels for close scenes and coarser ones for distant scenes is beneficial. LoD also resolves potential ``collision" issues (right side of Figure \ref{fig:lod_large}), where Gaussians from previous views overlap with those in subsequent frames in a long sequence, which will have a negative impact on camera pose tracking. By applying coarser details to distant Gaussians, LoD ensures that nearby reconstructions remain clear, enhancing efficiency and accuracy in large-scale outdoor mapping.

We voxelize the point cloud with varying voxel sizes based on camera distance, creating a hierarchical structure for rendering efficiency. The scene is divided into multiple levels, each with different resolutions, from fine to coarse. Given \( m \) levels of voxel sizes \( \{\epsilon_1, \cdots, \epsilon_m\} \) and LoD thresholds \( \{r_1, \cdots, r_{m-1}\} \), each voxel is assigned a specific level of detail \( L \in \mathbb{N} \), and sparse voxels within the field of view are selected based on distance. The voxelization process proceeds as follows:
\begingroup
\small
\begin{equation}
\mathbf{V} = \left\{ \left\lfloor \frac{\mathbf{P}}{\epsilon_l} \right\rfloor \cdot \epsilon_l \mid \epsilon_l \in \{\epsilon_1, \cdots, \epsilon_m\} \right\},
\label{eq:voxelize}
\end{equation}
\endgroup
where \( \mathbf{P}\in\mathbb{R}^3 \) is the point cloud position, and \( \epsilon_l \) corresponds to the voxel size at level \( l \), determined by the camera distance.  The voxelization process is similar to Octree-GS \cite{ren2024octree} but we do not maintain an octree.

When rendering, appropriate sparse voxels are then selected based on their proximity to the camera. To determine the voxel selection mask, we calculate the Euclidean distance $d_v = \|\mathbf{x}_v - \mathbf{x}_c\|_2$ between each voxel $\mathbf{x}_v$ and the camera $\mathbf{x}_c$. For each level $l$, voxels are selected if: 1) They fall within the distance range $[r_{l-1}, r_l)$; 2) Their level label $L(\mathbf{x}_v)$ matches the current level $l$; 3) They are visible from the camera. The mask is computed as:
\begin{equation}
\begin{aligned}
\text{mask}'_{i} &= 
\begin{cases} 
1, & d_v < r_1 \,\,\&\,\, L_i == 1, \\
1, & r_{l-1} \leq d_v < r_l \,\,\&\,\, L_i == l, \\
1, & d_v \geq r_{m-1} \,\,\&\,\, L_i == m, \\
0, & \text{otherwise}
\end{cases} \\
\text{mask}_{i} &= \text{mask}'_{i} \,\,\&\,\, \text{visible}(\mathbf{x}_v).
\end{aligned}
\end{equation}

This mask ensures that only voxels within the specified level distance range and visible to the camera are selected, which will be feed into MLPs to generate Gaussians for Splat operation mentioned in Section \ref{sec:splat}.

\paragraph{Map Update}

We optimize the map representation based on the method in \cite{kerbl20233gs}, using a combination of $L_1$ color distance and SSIM to constrain the current view. 
\begin{equation}
    \mathcal{L}_{\text{render}} = \sum_{m=1}^{M} \lambda_1 \Vert \mathbf{I}^{GS}_m - \mathbf{I}_m^{gt} \Vert_1 + \lambda_2 \, \text{SSIM}(\mathbf{I}^{GS}_m, \mathbf{I}_m^{gt}),
\end{equation}
where $m$ represents the pixel coordinates of the current image, $\mathbf{I}^{GS}$ is the rendered RGB image, $\mathbf{I}^{gt}$ is the ground truth RGB image, $\text{SSIM}(,)$ is the D-SSIM term, and $\lambda_1$ and $\lambda_2$ are hyperparameters.

To improve the geometric accuracy of 3D Gaussian depth rendering, we apply the smoothing method from \cite{chung2023depth} to reduce overfitting in sparse monocular input settings:
\begin{equation}
  \mathcal{L}_{\text{smooth}} = \sum_{d_j \in \text{Adj}(d_i)} \mathbf{1}_{ne}(d_i, d_j) |d_i - d_j|^2,
  \label{eq:smoothloss}
\end{equation}
where $\mathcal{L}_{\text{smooth}}$ denotes the smoothness loss, $\text{Adj}(d_i)$ represents the set of neighboring points to $d_i$, and $\mathbf{1}_{ne}(d_i, d_j)$ is an indicator function for whether $d_i$ and $d_j$ form a geometric edge. $|d_i - d_j|^2$ is the Euclidean distance between $d_i$ and $d_j$, with edges extracted using a Canny operator \cite{canny}.

We bring isotropic regularisation in MonoGS \cite{matsuki2024gaussian} to penalize primitives with a high aspect ratio:
\begingroup
\small
\begin{equation}
    \mathcal{L}_{\text{iso}} = \sum_{i=1}^{\vert G \vert} \Vert\mathbf{s}_{i}-\bar{\mathbf{s}}_i\cdot \mathbf{1}\Vert_1,
\end{equation}
\endgroup
where $\vert G \vert$ is the number of Gaussian primitives being optimized, and $\bar{\mathbf{s}}$ is the mean scaling.

The total loss function for optimization is defined as:
\begin{equation}
    \mathcal{L}_{\text{total}} = \mathcal{L}_{\text{render}} + \lambda_{\text{i}} \mathcal{L}_{\text{iso}} + \lambda_{\text{s}} \mathcal{L}_{\text{smooth}} ,
\end{equation}
where $\lambda_{\text{i}}$ and$ \lambda_{\text{s}}$ are hyperparameters.

\paragraph{Map Expansion}

A key challenge in large-scale dynamic expansion is the overlap of newly created voxels with existing ones. Due to the complex geometry of long outdoor sequences, current 3D GS SLAM methods struggle to efficiently detect such duplicates. A fast method is needed to check for voxel duplication, especially when expanding into unexplored areas and registering new maps. In our system, new voxels are generated by estimating the camera pose $\boldsymbol{\xi}$, transforming the metric depth into a point cloud, and voxelizing it hierarchically based on distance. Due to significant visual overlap between consecutive viewpoints, many newly created voxels already exist in the map, causing redundancy.

Our scene representation addresses this issue through a spatial hashing mechanism. Specifically, we deduplicate the anchor points of voxels by applying a spatial hash function \cite{spatialhash}:
\begin{equation}
h(x) = \left( \bigoplus_{i=1}^{d} x_i \pi_i \right) \mod T,
\label{eq:hash}
\end{equation}
where $\mathbf{x} = (x_1, x_2, x_3)$ is the 3D coordinate of the anchor point, $\pi_i$ are prime numbers (following the setting of \cite{muller2022instant}, $\pi_1 = 1, \pi_2 = 2654435761, \pi_3 = 805459861$), and $T = 2^{63}$. This spatial hashing method allows us to perform deduplication in constant time, ensuring that only unique voxels are retained in the map. By reducing redundant voxel entries, this approach is essential for efficient large-scale voxelized Gaussian Splatting SLAM, minimizing memory usage and computational overhead while preserving an accurate map in overlapping regions.

\subsection{Camera Tracking}

\paragraph{Online Pose Tracking}
We develop our tracking module based on DF-VO \cite{zhan2021df}, using optical flow differences for 2D-2D and 2D-3D tracking. Depth extraction is performed using networks like Monodepth \cite{monodepth}, DPT \cite{dpt}, and Depth Anything \cite{depthanything}, which work well for short sequences but suffer from metric ambiguity, degrading SLAM performance. UniDepth \cite{piccinelli2024unidepth} mitigates this ambiguity by standardizing the camera space transformation, allowing us to recover metric depth before tracking. To maintain depth consistency, we use RANSAC to correct scale errors between frames as described in \cite{zhan2021df}.

Feature points are extracted using DISK \cite{tyszkiewicz2020disk}, and the matched point pairs are used to estimate motion with LightGlue \cite{lindenberger2023lightglue}. These methods first establish 2D-2D correspondences, which are used to estimate camera motion via epipolar geometry \cite{zhan2021df}. Specifically, given a pair of images $(\mathbf{I}_i, \mathbf{I}_j)$, we could obtain a set of 2D-2D correspondences $(\mathbf{p}_i, \mathbf{p}_j)$. Using the epipolar constraint, the fundamental matrix $\mathbf{F}$ or the essential matrix $\mathbf{E}$ can be solved, where $\mathbf{F} = \mathbf{K}^{-T}\mathbf{E}\mathbf{K}^{-1}$ is related to the camera intrinsics $\mathbf{K}$. Decomposing $\mathbf{F}$ or $\mathbf{E}$ then allows us to recover the camera motion parameters $[\mathbf{R}, \mathbf{t}]$ of $\boldsymbol{\xi}$. If epipolar geometry fails due to motion degeneracy or scale ambiguity we employ the Geometric Robust Information Criterion (GRIC) \cite{zhan2021df} to select the appropriate motion model. In cases where the essential matrix is unreliable, we switch to the Perspective-n-Point (PnP) method, which estimates the camera pose by minimizing reprojection error using 2D-3D correspondences. Further details are provided in the Appendix.

\subsection{Loop Closure}
In our system, we integrate a proximity-based loop closure detection with a traditional SLAM back-end to enhance long-term localization accuracy. We detect loop closures using image retrieval techniques based on DBoW2 \cite{galvez2012bags}, followed by Sim(3) optimization \cite{dpv-slam, strasdat2010scale} with a smoothness term and loop closure constraints:

\begin{equation}
\begin{aligned}
\underset{S_1,\cdots,S_N}{\arg \min} \sum_{i}^{N} \Vert \log_{\text{Sim}(3)}(\Delta S_{i,i+1}^{-1}\cdot S_{i}^{-1}\cdot S_{i+1})\Vert^2_2 \\
+ \sum_{(j,k)}^L \Vert \log_{\text{Sim}(3)}(\Delta S_{jk}^{\text{loop}}\cdot S_{j}^{-1}\cdot S_{k})\Vert^2_2,
\end{aligned}
\end{equation}
where $S_i$ is the absolute similarity of keyframe $i$, the first term is the smoothness term between consecutive keyframes, the second term represents the error for the loop closure between keyframes $j$ and $k$, $\Delta S$ is the relative similarity between keyframes. Details could be found in the supplementary material.

After pose update, Our approach applies a rigid transformation to all voxels across different levels to maintain spatial consistency. We update all voxels in the global coordinate frame to align with the optimized camera poses. 
Let \( \mathbf{p}_i \in \mathbb{R}^3 \) be the \( i \)-th anchor point, originally associated with the \( j \)-th camera pose. 
Given the original pose \( \mathbf{T}_{\text{old}}^{(j)} \in \text{SE}(3) \) and the optimized pose \( \mathbf{T}_{\text{new}}^{(j)} \in \text{SE}(3) \), 
the updated position of the anchor point is given by:

\begin{equation}
\mathbf{p}_i^{\text{new}} = \mathbf{T}_{\text{new}}^{(j)} \cdot \left( \mathbf{T}_{\text{old}}^{(j)} \right)^{-1} \cdot 
\begin{bmatrix}
\mathbf{p}_i \\
1
\end{bmatrix}.
\label{eq:global_update}
\end{equation}

To reduce memory consumption for large-scale voxels, we process the update in batches. 
The entire anchor point set of voxels \( \mathcal{P} = \{ \mathbf{p}_1, \dots, \mathbf{p}_N \} \) is divided into \( M \) disjoint subsets \( \mathcal{B}_1, \dots, \mathcal{B}_M \), 
each of size \( B \ll N \). Within each batch \( \mathcal{B}_m \), the update rule remains consistent, but applied locally:

\begin{equation}
\forall\, \mathbf{p} \in \mathcal{B}_m,\quad
\mathbf{p}^{\text{new}} = \mathbf{T}_{\text{new}}^{(\pi(\mathbf{p}))} \cdot \left( \mathbf{T}_{\text{old}}^{(\pi(\mathbf{p}))} \right)^{-1} \cdot 
\begin{bmatrix}
\mathbf{p} \\
1
\end{bmatrix},
\label{eq:batch_update}
\end{equation}

where \( \pi(\mathbf{p}) \) maps each point \( \mathbf{p} \) to its associated camera pose index. This ensures that anchor points remain correctly aligned across the updated camera coordinate frames. Subsequently, a re-voxelization process (Eq. \ref{eq:voxelize}) is required to adaptively refine the map structure. To efficiently manage voxels introduced by loop closure corrections, we leverage the spatial hashing mechanism (Eq. \ref{eq:hash}), which enables fast lookup of updated voxels.

\section{Experiments}

\begin{figure*}[t]
    \centering
    \includegraphics[width=\linewidth]{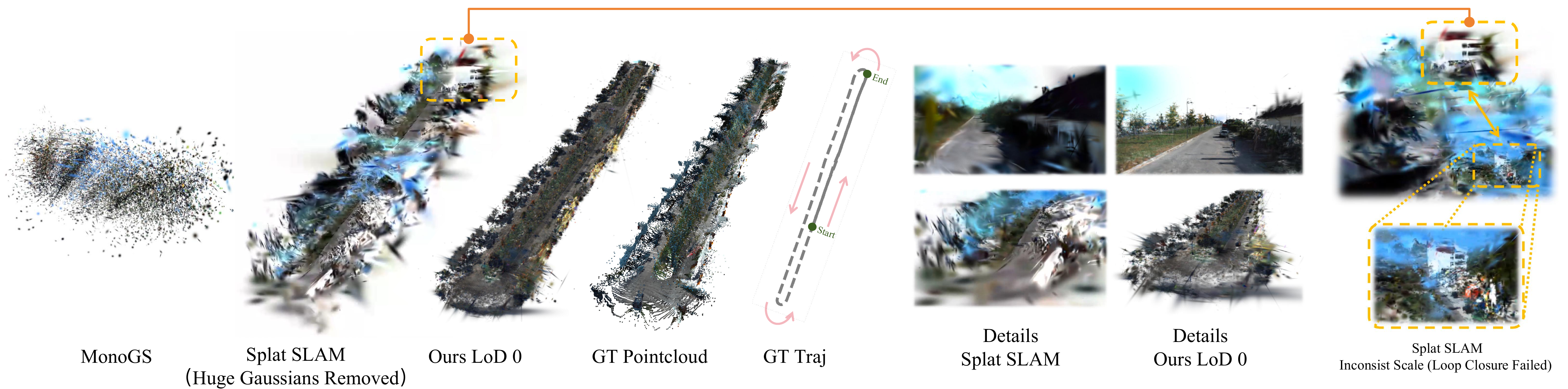}
    \caption{This figure evaluates rendering and geometric quality via global view on KITTI 06 Sequence. Splat SLAM exhibits severe scale inconsistency , whereas our method maintains precise scale coherence even in long outdoor sequences.}
    \label{fig:kitti_rendering_overall}
\end{figure*}

\begin{table*}[h]
  \centering
  \label{tab:table1}
  \centering
 \resizebox{0.85\linewidth}{!}{\textbf{}
    \begin{tabular}{l|c|c|c|ccccccccccc}
    \toprule
        \textbf{Methods} & \textbf{LC} & \textbf{Render} & \textbf{Avg.} & \textbf{00} & \textbf{01} & \textbf{02} & \textbf{03} & \textbf{04} & \textbf{05} & \textbf{06} & \textbf{07} & \textbf{08} & \textbf{09} & \textbf{10}   \\ \midrule

        \textitgray{seq. frames} & \textitgray{-} & \textitgray{-}&	\textitgray{2109} &	\textitgray{4542}&	\textitgray{1101}&	\textitgray{4661}&	\textitgray{801}&	\textitgray{271}&	\textitgray{2761}&	\textitgray{1101}&	\textitgray{1101}&	\textitgray{4071}&	\textitgray{1591}&	\textitgray{1201}\\
        
        \textitgray{seq. length (m)} & \textitgray{-} & \textitgray{-} & \textitgray{2012.243} & \textitgray{3724.19} & \textitgray{2453.20} & \textitgray{5067.23} & \textitgray{560.89} & \textitgray{393.65} & \textitgray{2205.58} & \textitgray{1232.88} & \textitgray{649.70} & \textitgray{3222.80} & \textitgray{1705.05} & \textitgray{919.52} \\
        
        \textitgray{contains loop} & \textitgray{-} & \textitgray{-} & \textitgray{-} & \textitgray{\ding{51}} & \textitgray{\ding{55}} & \textitgray{\ding{51}} & \textitgray{\ding{55}} & \textitgray{\ding{55}} & \textitgray{\ding{51}} & \textitgray{\ding{51}} & \textitgray{\ding{51}} & \textitgray{\ding{51}} & \textitgray{\ding{51}} & \textitgray{\ding{55}} \\
        
        \midrule
        ORB-SLAM2 (w/o LC) \cite{orbslam2} & \redtext{\ding{55}} & \redtext{\ding{55}}  & 69.727 & 40.65  & 502.20  & 47.82  & \textbf{0.94}  & 1.30  & 29.95  & 40.82  & 16.04  & 43.09  & 38.77  & 5.42    \\ 
        
        ORB-SLAM2 (w/ LC) \cite{orbslam2} & \greentext{\ding{51}} & \redtext{\ding{55}}  & 54.816 & \textbf{6.03}  & 508.34  & 14.76  & 1.02  & 1.57  & 4.04  & 11.16  & 2.19  & 38.85  & 8.39  & 6.63   \\ 
        
        LDSO \cite{ldso} & \greentext{\ding{51}} & \redtext{\ding{55}}  & 22.425 & 9.32  & 11.68  & 31.98  & 2.85  & 1.22  & 5.10  & 13.55  & 2.96  & 129.02  & 21.64  & 17.36   \\ 
        
        DF-VO \cite{zhan2021df} & \redtext{\ding{55}} & \redtext{\ding{55}}  & 16.440 & 14.45 & 117.40 & 19.69 &  1.00 & 1.39  &  \textbf{3.61} & 3.20  & \textbf{0.98}  & 7.63  & 8.36 &  3.13  \\ 
        
        DROID-VO \cite{droidslam} & \redtext{\ding{55}} & \redtext{\ding{55}}  & 54.188 & 98.43  & 84.20  & 108.80  & 2.58  & 0.93  & 59.27  & 64.40  & 24.20  & 64.55  & 71.80  & 16.91   \\ 
        
        DPVO \cite{dpvo}& \redtext{\ding{55}} & \redtext{\ding{55}}  & 53.609 & 113.21  & 12.69  & 123.40  & 2.09  & \textbf{0.68}  & 58.96  & 54.78  & 19.26  & 115.90  & 75.10  & 13.63     \\ 
        
        DROID-SLAM \cite{droidslam} & - & \redtext{\ding{55}}  & 100.278 & 92.10  & 344.60  & 107.61 & 2.38  & 1.00  & 118.50  & 62.47  & 21.78  & 161.60  & 72.32 & 118.70   \\ 
        
        DPV-SLAM \cite{dpv-slam} & \greentext{\ding{51}} & \redtext{\ding{55}}  & 53.034 & 112.80  & \textbf{11.50}  & 123.53  & 2.50  & 0.81  & 57.80  & 54.86  & 18.77  & 110.49  & 76.66  & 13.65   \\
        
        DPV-SLAM++ \cite{dpv-slam} & \greentext{\ding{51}} & \redtext{\ding{55}}  & 25.749 & 8.30  & 11.86  & 39.64  & 2.50  & 0.78  & 5.74  & 11.60  & 1.52  & 110.90  & 76.70  & 13.70    \\ 
        
        MonoGS \cite{matsuki2024gaussian} & \redtext{\ding{55}} & \greentext{\ding{51}}  & / & \textitgray{failed} & 543.47 & \textitgray{failed} & \textitgray{failed} & 20.75 & \textitgray{failed} & 137.22 & \textitgray{failed} & \textitgray{failed} & \textitgray{failed} & \textitgray{failed}  \\ 
        
        Splat-SLAM \cite{sandstrom2024splat} & - & \greentext{\ding{51}}  & / & \textitgray{83.07}$^\times$ & \textitgray{failed}  & \textitgray{failed} & 3.40 & 1.72 & \textitgray{33.01}$^\times$ & 130.75 & 14.35 & \textitgray{52.07}$^\times$ & \textitgray{27.42}$^\times$ &  63.55  \\ 
        
        \midrule
        Ours (w/o LC) & \redtext{\ding{55}} & \greentext{\ding{51}} &	16.437  & 7.09 &	129.74 & 	12.34 & 	2.49& 	2.25 &	5.92 & 	2.61 & 	2.59 & 	9.48 &	4.03 & 	2.27   \\ 
        
        Ours (w/ LC) & \greentext{\ding{51}} & \greentext{\ding{51}} &	\textbf{15.576}  & 6.83 & 	127.39 &	\textbf{11.30} &	2.18 &	1.88 &	4.36 &	\textbf{2.11} &	2.12 &	\textbf{7.04} &	\textbf{3.94} &	\textbf{2.18}     \\ \bottomrule
    \end{tabular}
  }
  \caption{Camera Tracking Results (ATE RMSE [m] $\downarrow$) on the KITTI Dataset. \textit{LC} denotes \textit{loop closure}. DROID-SLAM and Splat-SLAM use implicit loop detection via the pose factor graph, which works indoors but fails in large outdoor environments (see suppl.). \textitgray{[num]}$^\times$ indicates that Splat-SLAM crashes in Mapping Mode, where values are obtained with Tracking-only Mode. Our method is the only approach capable of achieving high-fidelity rendering from the current viewpoint while maintaining relatively strong tracking performance on long-sequence outdoor datasets. }
  \label{table:kitti_ate}
\end{table*}

\subsection{Experimental Setup}

We designed our experimental setup to evaluate GigaSLAM's scalability and versatility across diverse environments by using large outdoor datasets, with comprehensive metrics assessing both tracking accuracy and map quality.

\subsubsection{Dataset and Metrics}
We evaluate our system on datasets: KITTI \cite{geiger2012kitti}, KITTI 360 \cite{liao2022kitti} 4 Seasons\cite{wenzel20214seasons}, A2D2\cite{geyer2020a2d2}. KITTI is the primary dataset due to its kilometer-scale, long sequences, offering a challenging outdoor environment for SLAM. Unlike other methods, which are limited to smaller datasets or indoor scenes, our approach effectively handles large-scale outdoor scenarios.

For tracking accuracy, we report the Absolute Trajectory Error (ATE) \cite{ate} of the keyframes on three dataset: KITTI \cite{geiger2012kitti}, KITTI 360 \cite{liao2022kitti} 4 Seasons\cite{wenzel20214seasons}. Due to the absence of GT pose matrices in the A2D2 dataset, ATE computation becomes infeasible. We thus visualize our algorithm's tracking performance through projecting of its trajectory on Google Map in Figure \ref{fig:kitti_360_cmp}. Mapping quality is evaluated using photometric rendering metrics such as Peak Signal-to-Noise Ratio (PSNR) \cite{psnrssim}, Structural Similarity Index Measure (SSIM) \cite{psnrssim}, and Learned Perceptual Image Patch Similarity (LPIPS) \cite{lpips}, which collectively capture both pixel-level and perceptual differences in rendered images.

\subsubsection{Implementation Details}
Our experiments were conducted on machine with Ubuntu 22.04, and equipped with 12 Intel Xeon Gold 6128 3.40 GHz CPUs, 67GB of RAM, and an NVIDIA RTX 4090 GPU with 24GB of VRAM for the majority of tests. For certain ultra-long sequences, such as those in KITTI and KITTI-360 that exceed 4,000 frames, we utilized a high-memory system with 128GB of RAM, 20 Intel Xeon Platinum 8467C CPUs, and an NVIDIA L20 GPU with 48GB of VRAM to accommodate large-scale outdoor scenes. Our SLAM pipeline builds on the code structure of MonoGS in PyTorch, leveraging CUDA to accelerate splatting operations. To ensure runtime efficiency, we use a multi-process setup for tracking, mapping and loop closure.

Given the detail richness of outdoor scenes, we use a rendering resolution of 480 pixels in width (scaled proportionally in height) to optimize computational efficiency and memory usage. To evaluate reconstruction quality, these rendering images are upsampled to the original resolution using bicubic interpolation for efficiency. Employing a deep learning-based super-resolution algorithm, however, may yield higher reconstruction quality than the values reported in this section.

\subsection{KITTI Dataset}

\begin{figure}[t]
    \centering
    \includegraphics[width=\linewidth]{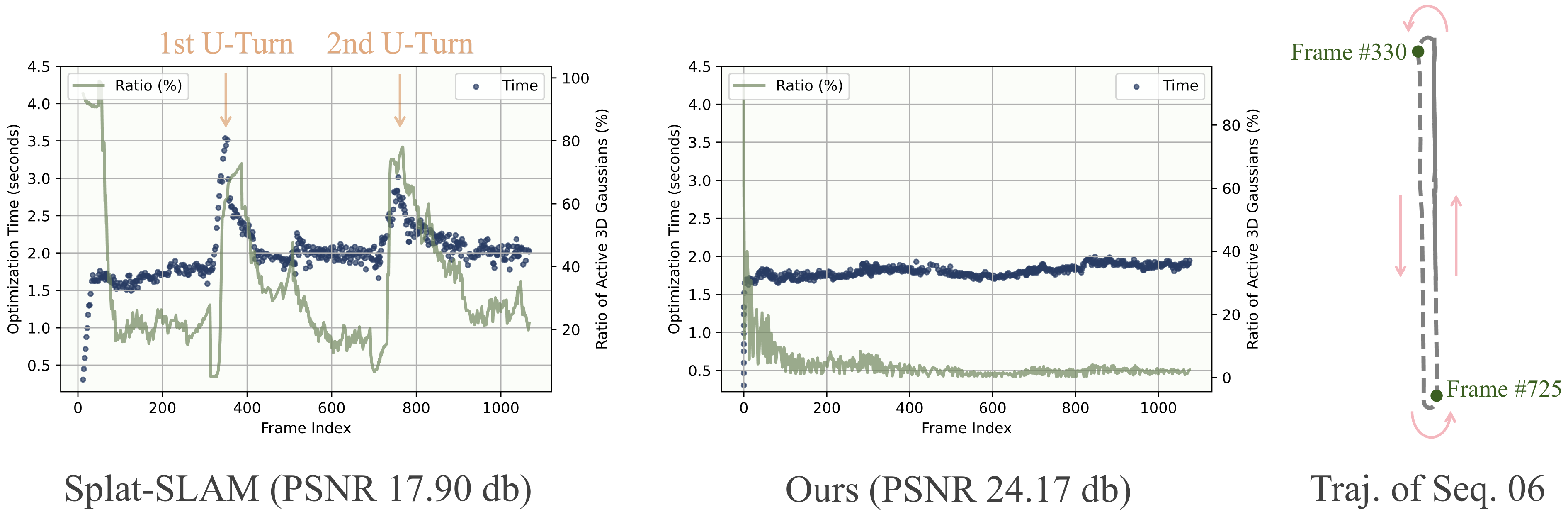}
    \caption{Comparison of rendering efficiency between Splat-SLAM and our method on KITTI-06. Splat-SLAM suffers efficiency drops at U-turns due to excessive visible Gaussians, affecting distant detail, while our method remains stable.}
    \label{fig:runtime_cmp}
\end{figure}

\begin{figure*}[htbp]
    \centering
    \begin{subfigure}{0.49 \textwidth}
        \centering
        \includegraphics[width=\linewidth]{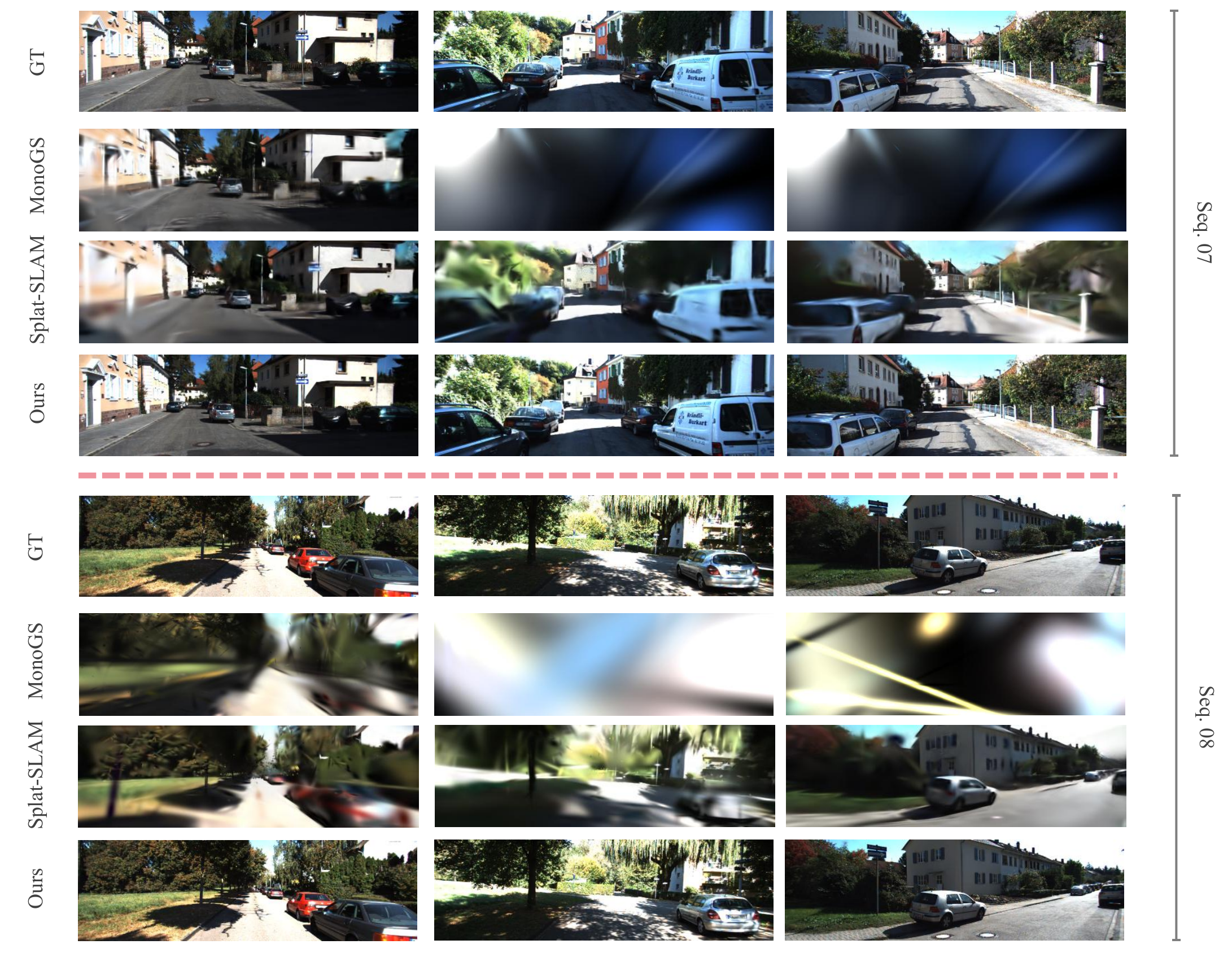}
        \caption{Sequence 07 \& 08}
    \end{subfigure}
    \begin{subfigure}{0.49 \textwidth}
        \centering 
        \includegraphics[width=\linewidth]{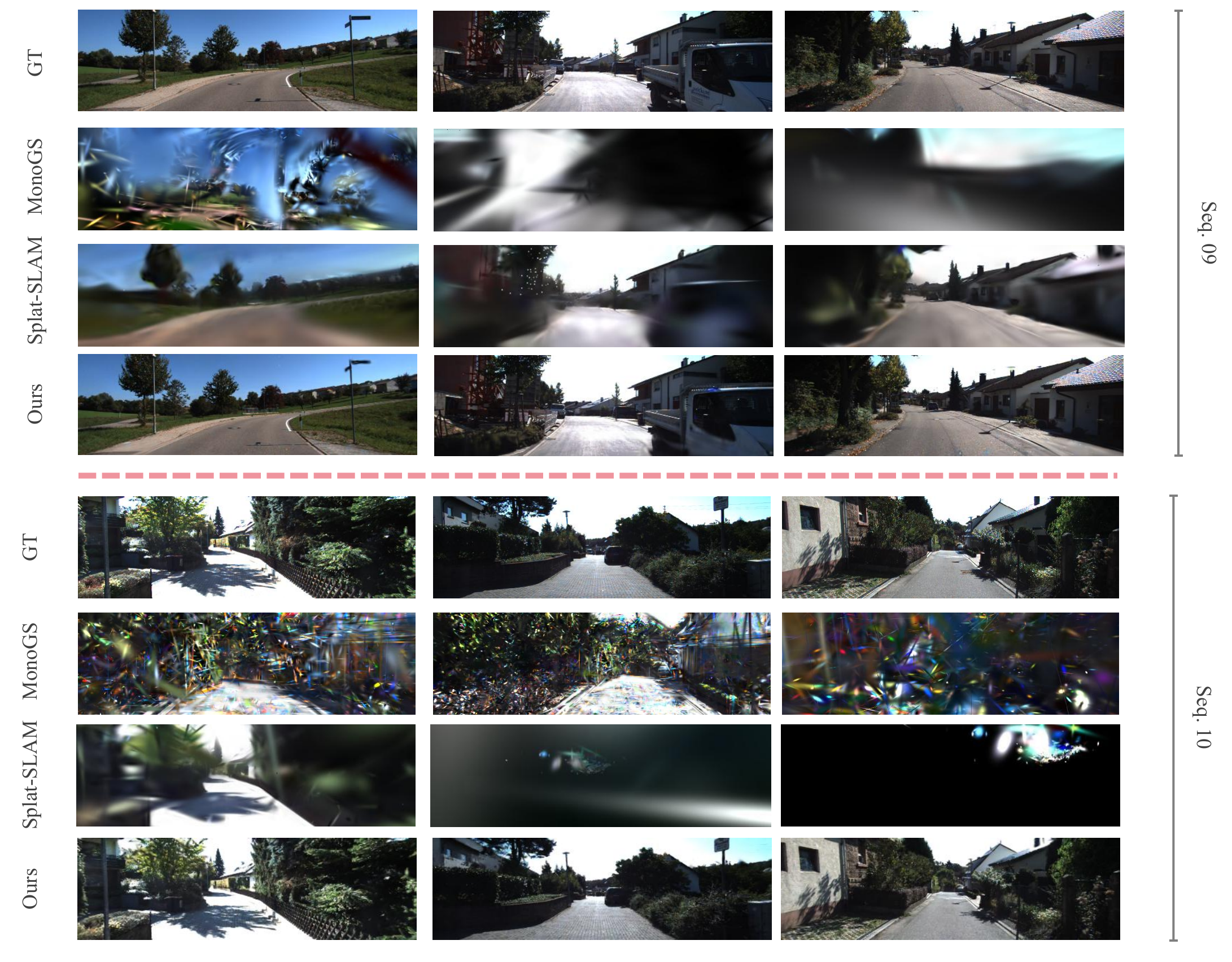}
        \caption{Sequence 09 \& 10}
    \end{subfigure}

    \caption{Rendering results on the KITTI dataset are shown for MonoGS (with RGB input) Splat-SLAM and our proposed method. Our method maintains stable rendering across extended outdoor sequences, while MonoGS and Splat-SLAM struggle on this scene.
    }
    \label{fig:kitti_rendering}
\end{figure*}

\begin{figure}[t]
  \centering
   \includegraphics[width=\linewidth]{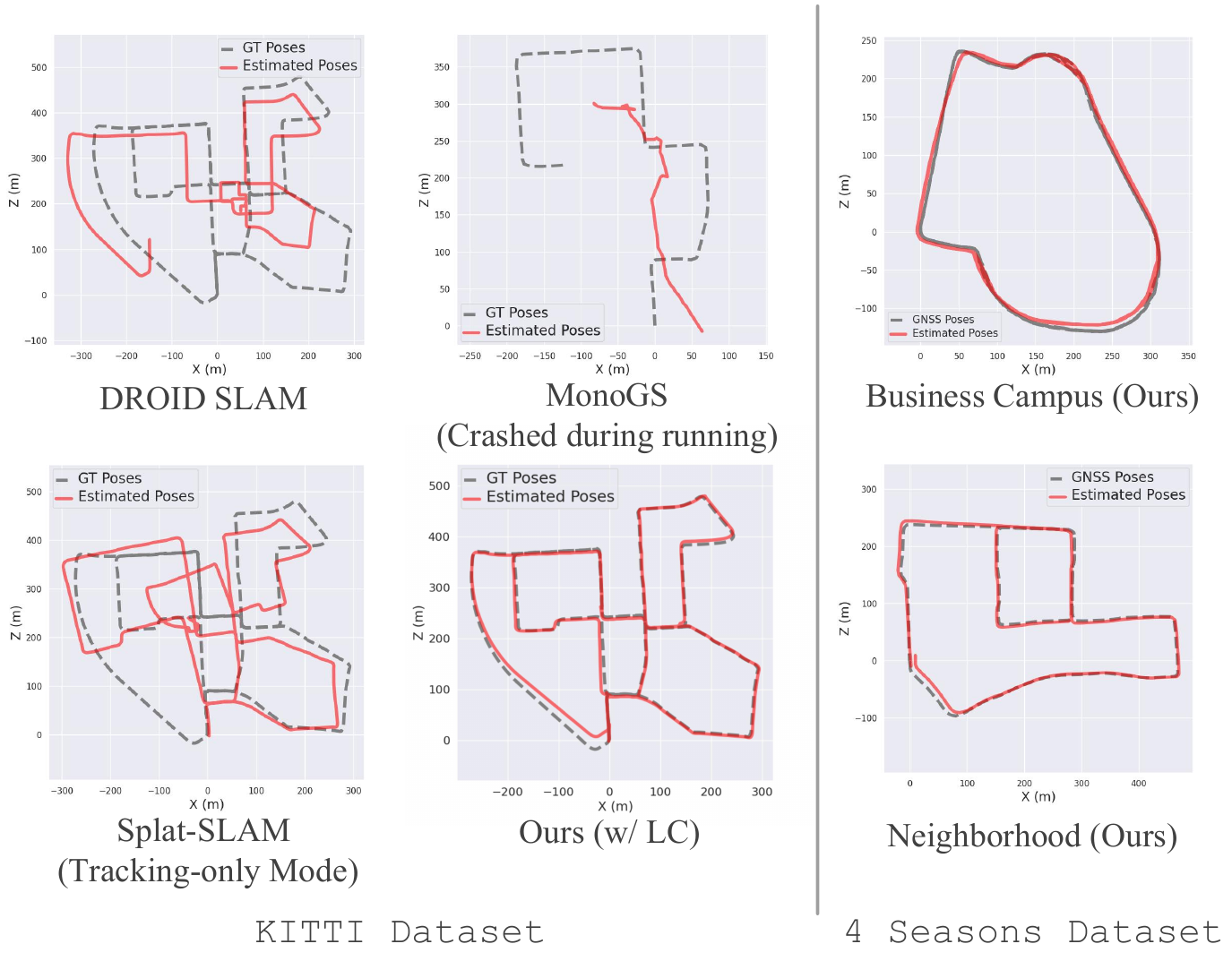}
   \caption{(Left) Trajectory estimation of different SLAM methods on sequence 00 of the KITTI dataset with RGB input. Our method demonstrates stable tracking over long outdoor scenes, unlike the scale drift in DROID-SLAM and the tracking failure of MonoGS. (Right) Our method on 4 Seasons dataset.}
   \label{fig:kitti_cmp}
\end{figure}

Table \ref{table:kitti_ate} presents the tracking performance of our method on the KITTI \cite{geiger2012kitti} dataset (data from DPV-SLAM \cite{dpv-slam}). Overall, our method shows strong tracking performance, particularly on long and complex sequences. Figure \ref{fig:kitti_cmp} compares camera trajectories on sequence 00, where our approach maintains stable and accurate pose estimates. In contrast, DROID-SLAM \cite{droidslam} exhibits significant scale drift on long sequences, indicating limited robustness. MonoGS \cite{matsuki2024gaussian} performs worse—it crashes after a few hundred frames and fails to continue tracking. As shown in Figure \ref{fig:kitti_rendering} and Table \ref{table:kitti_render}, the inaccurate poses from MonoGS degrade both mapping and tracking quality, making it unsuitable for KITTI’s long, outdoor sequences. Our method, however, delivers consistently accurate poses over extended trajectories, showing robustness unmatched by these baselines.

Figure \ref{fig:kitti_rendering_overall} qualitatively compares reconstructions from our method (Ours LoD 0), MonoGS, and Splat SLAM on KITTI 06. From left to right, it shows global reconstruction, ground truth pointcloud and trajectory, and zoom-ins on local geometry. MonoGS outputs sparse, noisy maps with major detail loss. Splat SLAM captures more structure but suffers from scale inconsistency due to failed loop closure. Our method reconstructs more consistent geometry throughout the sequence and better preserves scene details, even in distant areas. Zoom-in views highlight our strength in capturing thin structures and avoiding over-splatting.

\begin{table*}[t]
  \centering
  \footnotesize
  \setlength{\tabcolsep}{3.5pt}          
  \renewcommand{\arraystretch}{1.05}     
  \resizebox{\linewidth}{!}{
    \begin{tabular}{l|c|ccccccccc|c|ccccc}
    \toprule
     & \multicolumn{10}{c}{\textbf{KITTI 360 Dataset}} &  \multicolumn{6}{|c}{\textbf{4 Seasons Dataset}}\\
    \midrule
         \textbf{Methods} & \textbf{Avg.}  & \textbf{0000} & \textbf{0002} & \textbf{0003} & \textbf{0004} & \textbf{0005} & \textbf{0006} & \textbf{0007} & \textbf{0009} & \textbf{0010} & \textbf{Avg.}  & \textbf{Business Campus} & \textbf{Office Loop} & \textbf{Old Town} & \textbf{Neighborhood} & \textbf{City Loop} \\ \midrule
        
        \textitgray{seq. frames}  & \textitgray{8497} & \textitgray{11518}  & \textitgray{14607}  & \textitgray{1031}  & \textitgray{11587}  & \textitgray{6743}  & \textitgray{9699}  & \textitgray{3396}  & \textitgray{14056}  & \textitgray{3836}  & \textitgray{20200}  & \textitgray{17280}  & \textitgray{15177}  & \textitgray{28999}  & \textitgray{11121}  & \textitgray{28424}   \\ 
        
        \textitgray{seq. length (m)}  & \textitgray{6971.088}  & \textitgray{8403.15}  & \textitgray{11501.28}  & \textitgray{1378.73}  & \textitgray{9975.28}  & \textitgray{4690.74}  & \textitgray{7979.92}  & \textitgray{4887.51}  & \textitgray{10579.68}  & \textitgray{3343.49}  & \textitgray{4906.55}  & \textitgray{3132.38}  & \textitgray{3710.66}  & \textitgray{5258.60}  & \textitgray{2078.37}  & \textitgray{10352.74}  \\ 
        
        \textitgray{contains loop} & \textitgray{-} & \textitgray{\ding{51}} & \textitgray{\ding{51}} & \textitgray{\ding{55}} & \textitgray{\ding{51}} & \textitgray{\ding{51}} & \textitgray{\ding{51}} & \textitgray{\ding{55}} & \textitgray{\ding{51}} & \textitgray{\ding{55}} & \textitgray{-} & \textitgray{\ding{51}} & \textitgray{\ding{51}} & \textitgray{\ding{81}} & \textitgray{\ding{51}} & \textitgray{\ding{81}}\\
        
        \midrule

        DROID-SLAM \cite{droidslam}  & 193.307 & 110.07  & 233.87  & \textbf{10.79}  & 169.11  & 139.02  & 113.81  & 577.39  & 165.34  & 220.36 & / & OOM & 175.63 & OOM & 158.19 & OOM   \\ 
        
        Ours (w/o LC) &	61.402   & 25.53& 	56.59 &	29.13 &	116.15& 	20.51& 	28.52 &	203.02 &	16.83 &	56.35 & 99.098 & 8.81 & 36.85 & 69.71 & 7.48 & 372.64\\ 
        
        Ours (w/ LC) &	\textbf{47.107} & \textbf{17.72} &	\textbf{34.98} &	27.55 &	\textbf{40.54} &	\textbf{20.46} &	\textbf{19.12} &	\textbf{197.33} 	& \textbf{13.22} &	\textbf{53.03} &	\textbf{92.950} &	\textbf{7.33} &	\textbf{21.12} &	\textbf{67.89} &	\textbf{6.28} &	\textbf{362.13}  \\ 
        
        \bottomrule
    \end{tabular}
    }
    \caption{Camera Tracking Results (ATE RMSE [m] $\downarrow$) on the KITTI 360 \& 4 Seasons Dataset. OOM stands for Out-of-Memory.  \textitgray{\ding{81}} marks a closed-loop sequence with just one hardly detectable start-end loop point; DBoW failed to identify this latent loop closure.}
  \label{tab:kitti360}
  \vspace{-5pt}
\end{table*}

\begin{table}[t]
    \centering
 \resizebox{1.0\linewidth}{!}{
    \begin{tabular}{c|c|c|ccccccccccc}
    \toprule
        \textbf{Methods} & \textbf{Metrics}& \textbf{Avg.} & \textbf{00} & \textbf{01} & \textbf{02} & \textbf{03} & \textbf{04} & \textbf{05} & \textbf{06} & \textbf{07} & \textbf{08} & \textbf{09} & \textbf{10}   \\ \midrule
        MonoGS & PSNR $\uparrow$ & \textitgray{11.09} & \textitgray{10.09}$^\times$ & 16.40 & \textitgray{8.78}$^\times$ & \textitgray{11.83}$^\times$ & 17.43 & \textitgray{10.83}$^\times$ & 12.66 & \textitgray{8.69}$^\times$ & \textitgray{6.90}$^\times$ & \textitgray{9.71}$^\times$ & \textitgray{8.66}$^\times$  \\ 
        (RGB) & SSIM $\uparrow$ & \textitgray{0.38} & \textitgray{0.43}$^\times$ & 0.58 & \textitgray{0.31}$^\times$ & \textitgray{0.36}$^\times$ & 0.55 & \textitgray{0.37}$^\times$ & 0.43 & \textitgray{0.34}$^\times$ & \textitgray{0.27}$^\times$ & \textitgray{0.27}$^\times$ & \textitgray{0.32}$^\times$  \\ 
         \cite{matsuki2024gaussian} & LPIPS $\downarrow$ & \textitgray{0.79} & \textitgray{0.82}$^\times$ & 0.66 & \textitgray{0.85}$^\times$ & \textitgray{0.82}$^\times$ & 0.55 & \textitgray{0.84}$^\times$ & 0.76 & \textitgray{0.86}$^\times$ & \textitgray{0.89}$^\times$ & \textitgray{0.81}$^\times$ & \textitgray{0.84}$^\times$  \\ 
        \midrule
        Splat-SLAM & PSNR $\uparrow$ & / & \textitgray{20.27}$^\times$ & \textitgray{failed} & \textitgray{failed} & 21.10 & 19.42 & \textitgray{20.33}$^\times$ & 17.90 & 20.72 & \textitgray{20.48}$^\times$ & \textitgray{20.86}$^\times$ & 12.48\\ 
        (RGB) & SSIM $\uparrow$ & / & \textitgray{0.77}$^\times$ & \textitgray{failed} & \textitgray{failed} & 0.64 & 0.68 & \textitgray{0.67}$^\times$ & 0.61 & 0.70 & \textitgray{0.65}$^\times$ & \textitgray{0.68}$^\times$ & 0.31  \\ 
         \cite{sandstrom2024splat} & LPIPS $\downarrow$ & / & \textitgray{0.41}$^\times$ & \textitgray{failed} & \textitgray{failed} & 0.59 & 0.52 & \textitgray{0.55}$^\times$ & 0.66 & 0.51 & \textitgray{0.59}$^\times$ & \textitgray{0.58}$^\times$ & 0.76   \\ 
        \midrule
        Ours & PSNR $\uparrow$ & \textbf{24.22} & \textbf{24.14} & \textbf{24.91} & \textbf{22.71} & \textbf{24.40} & \textbf{25.22} & \textbf{24.92} & \textbf{24.17} & \textbf{24.88} & \textbf{23.42} & \textbf{23.03} & \textbf{24.09}  \\ 
        (RGB) & SSIM $\uparrow$ & \textbf{0.95} & \textbf{0.96} & \textbf{0.96} & \textbf{0.95} & \textbf{0.95} & \textbf{0.96} & \textbf{0.96} & \textbf{0.96} & \textbf{0.97} & \textbf{0.94} & \textbf{0.95} & \textbf{0.95}  \\ 
        ~ & LPIPS $\downarrow$ & \textbf{0.31} & \textbf{0.28} & \textbf{0.33} & \textbf{0.33} & \textbf{0.33} & \textbf{0.30} & \textbf{0.28} & \textbf{0.30} & \textbf{0.25} & \textbf{0.29} & \textbf{0.34} & \textbf{0.35} \\ \bottomrule
    \end{tabular}
      }
  \caption{Rendering performance on KITTI dataset. \textitgray{[num]}$^\times$ indicates that MonoGS or Splat-SLAM crashes before completing all frames, and the values are averaged over the processed frames before failure. \textitgray{failed} indicates that the tracking module of Splat-SLAM returned NaN values, causing the algorithm to fail for the entire sequence.}
  \label{table:kitti_render}
\end{table}

\subsection{KITTI 360, 4 Seasons \& A2D2 Dataset}

\begin{figure*}[t]
    \centering

    \includegraphics[width=0.85\linewidth]{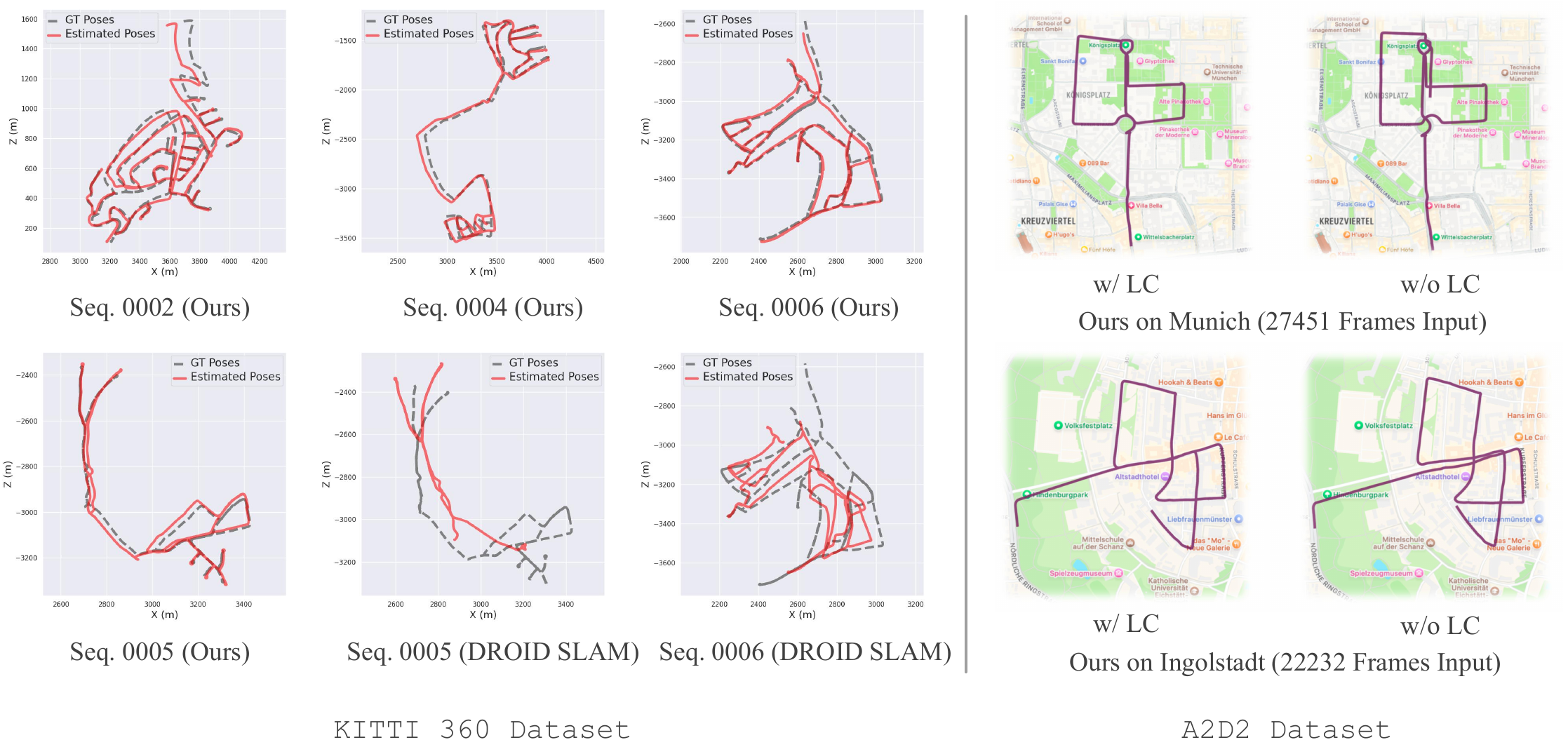}
    
    \caption{(Left) Camera trajectory on the KITTI 360 dataset. Unlike DROID SLAM, which fails due to implicit loop closure, our method explicitly handles large-scale trajectories. (Right) Our method on A2D2 dataset.}
    \label{fig:kitti_360_cmp}
\end{figure*}

To further evaluate the robustness of our method in extensive outdoor scenarios, we conducted an experiment on the KITTI 360 \cite{liao2022kitti}, 4 Seasons\cite{wenzel20214seasons} \& A2D2\cite{geyer2020a2d2} dataset. Unlike the KITTI \cite{geiger2012kitti} dataset, where sequence lengths peak at 4,661 frames with an average of approximately 2,109 frames, KITTI 360 sequences are significantly longer, averaging 8,497 frames and reaching a maximum length of 14,607 frames. So do 4 Seasons and A2D2 datasets that have an average of 20,000 frame input. These extended trajectories introduce unique challenges in monocular RGB-based SLAM, particularly due to the accumulation of errors over such long sequences, which can severely impact tracking accuracy and mapping fidelity. 

Notably, almost no existing monocular RGB-based SLAM or VO system has been fully evaluated on these datasets to date. Our method, however, demonstrated the capability to process these ultra-long sequences effectively, providing stable and continuous camera pose estimations across the full length of each sequence. Initial results (Figure \ref{fig:kitti_360_cmp} and Table \ref{tab:kitti360}) indicate acceptable  performance on ultra-long sequences, highlighting our system’s resilience in mitigating error accumulation over extended trajectories. On the 4 Seasons dataset, the substantial increase in the number of input frames leads to a dramatic rise in memory requirements, rendering DROID-SLAM inapplicable due to GPU memory exhaustion. As shown, DROID-SLAM runs out of memory on most sequences. Even in the two sequences where it is able to run, our method significantly outperforms it. Although our approach demonstrates strong performance on these sequences, potential loop closures were not detected due to dataset limitations (i.e., DBoW failed to find matches). Nevertheless, our system still maintains reasonable tracking performance under such challenging conditions. This underscores the scalability and robustness of our approach for unbounded, long-sequence outdoor SLAM tasks.

\subsection{Ablation and further Studies}

We performed ablation studies on KITTI seq. 06 (Table \ref{tab:ablation}), incorporating a depth prior into MonoGS (RGB) and adding the LoD GS module to its backend. Despite the depth prior, MonoGS shows performance degradation (Section \ref{sec:mapping}). Integrating LoD GS partially mitigates this, and with additional modules, our method outperforms others. LoD GS improves rendering by adaptively selecting voxel sizes based on distance, balancing detail and efficiency (Table \ref{tab:voxel_hierarchy}).

\setlength{\intextsep}{2pt}   
\setlength{\columnsep}{6pt}    
\begin{wraptable}{r}{0.4\linewidth}  
\centering
\renewcommand{\arraystretch}{0.8}  
\setlength{\tabcolsep}{2pt}  
\resizebox{1.05\linewidth}{!}{  
\begin{tabular}{lc}
\toprule
\textbf{Method} & \textbf{ATE [m]} \\ \midrule
MonoGS & 137.22 \\ 
MonoGS + UniDepth & 100.03 \\ 
Ours w/ LoD GS only & 47.33 \\ 
Ours w/o LC & 2.61 \\ 
Ours w/ LC & 2.11 \\ 
\bottomrule
\end{tabular}
}
\caption{Ablation experiments of components}
\label{tab:ablation}
\end{wraptable}

We compared our method with Splat-SLAM \cite{sandstrom2024splat} on the KITTI dataset and observed that while Splat-SLAM  performs well in indoor environments, its rendering time increases significantly with the number of frames in large outdoor sequences. In contrast, our method maintains more stable performance, as its rendering time does not scale as drastically. To ensure a fair comparison, both methods were tested with the same rendering resolution, 3DGS CUDA rasterization code, and identical optimization settings. As shown in Figure \ref{fig:runtime_cmp}, in the KITTI-06 sequence, Splat-SLAM’s active 3D Gaussian ratio spikes to nearly $80\%$ after two U-turns, leading to unstable computation. Our method, leveraging a hierarchical voxelized 3D Gaussian representation, effectively controls the number of Gaussians involved in optimization, ensuring more stable and efficient performance in large-scale outdoor environments.

\begin{table}[h]
    \centering
    \resizebox{\linewidth}{!}{ 
        \begin{tabular}{cllccc}
            \toprule
            Level(s) Num. & Voxel Size & Distance Partition & Avg. Frustum Vox. Num. & PSNR & GPU MEM \\
            \midrule
            1 level      & [0.1]                      & []                  & 411,612 & 24.29 db & 22.46 GiB\\
            2 levels     & [0.1, 0.25]               & [20]              & 223,627 & 24.64 db & 15.82 GiB\\
            3 levels     & [0.1, 0.25, 1]            & [20, 40]  & 116,639 & 24.05 db & 11.77 GiB \\
            4 levels     & [0.1, 0.25, 1, 5]         & [20, 40, 80]       & 34,721 & 24.21 db & 9.46 GiB \\
            5 levels     & [0.1, 0.25, 1, 5, 25]     & [20, 40, 80, 160]  & 21,342 & 24.17 db & 8.62 GiB \\
            \bottomrule
        \end{tabular}
    }
    \caption{Ablation study of LoD on KITTI 06 Seq.}
    \label{tab:voxel_hierarchy}
\end{table}
\section{Conclusion}

We present GigaSLAM, the first SLAM system for long-term, kilometer-scale outdoor sequences using monocular RGB input. By employing a hierarchical sparse voxel structure and a metric depth module, GigaSLAM enables efficient large-scale mapping and robust pose estimation. Evaluated on the KITTI, KITTI-360, 4 Seasons and A2D2 datasets, our system demonstrates strong performance and good scalability for outdoor SLAM tasks. Looking ahead, future work will focus on improving loop closure detection, particularly under high-speed motion. Enhancing system stability in such scenarios will be key to advancing reliable SLAM for ultra-large-scale, real-world applications. Moreover, extending GigaSLAM to operate under more dynamic conditions and diverse outdoor environments could further improve its practicality.
{
    \small
    \bibliographystyle{ieeenat_fullname}
    \bibliography{main}
}

\clearpage
\appendix

\clearpage
\setcounter{page}{1}
\setcounter{section}{0} 
\maketitlesupplementary

\renewcommand{\thesection}{\Alph{section}}

\renewcommand{\thefigure}{\thesection.\arabic{figure}}
\renewcommand{\thetable}{\thesection.\arabic{table}}
\renewcommand{\theequation}{\thesection.\arabic{equation}}

\makeatletter
\@addtoreset{figure}{section}
\@addtoreset{table}{section}
\@addtoreset{equation}{section}
\makeatother

\section{What Challenges Are We Facing on Outdoor Long-Sequence Datasets?}
\label{sec:challenges}

\begin{figure*}[h]
    \centering

    \begin{subfigure}{0.19 \linewidth}
        \centering
        \includegraphics[width=\linewidth]{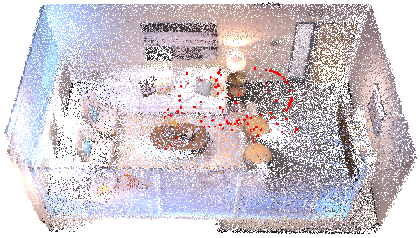}
        \caption{Indoor Scene}
    \end{subfigure}
    \begin{subfigure}{0.19 \linewidth}
        \centering
        \includegraphics[width=\linewidth]{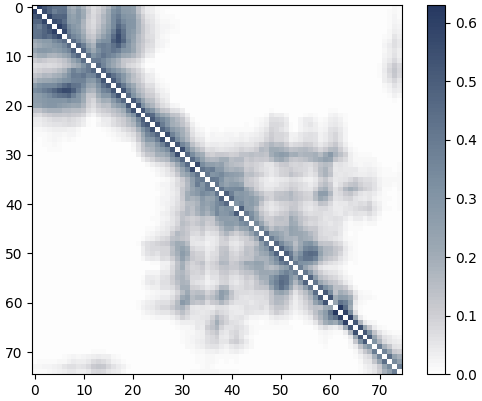}
        \caption{Replica Room 0}
    \end{subfigure}
    \begin{subfigure}{0.19 \linewidth}
        \centering
        \includegraphics[width=\linewidth]{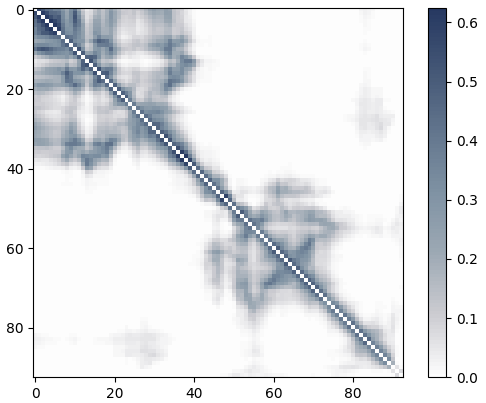}
        \caption{Replica Room 2}
    \end{subfigure}
    \begin{subfigure}{0.19 \linewidth}
        \centering
        \includegraphics[width=\linewidth]{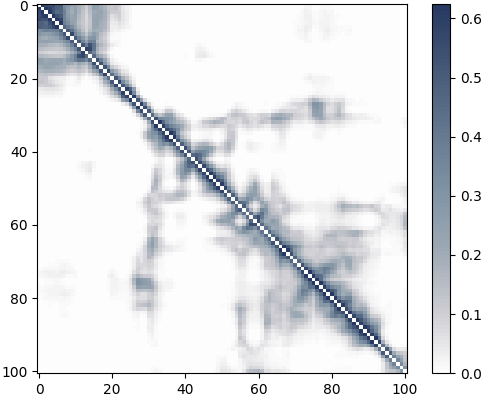}
        \caption{Replica Office 2}
    \end{subfigure}
    \begin{subfigure}{0.19 \linewidth}
        \centering
        \includegraphics[width=0.97\linewidth]{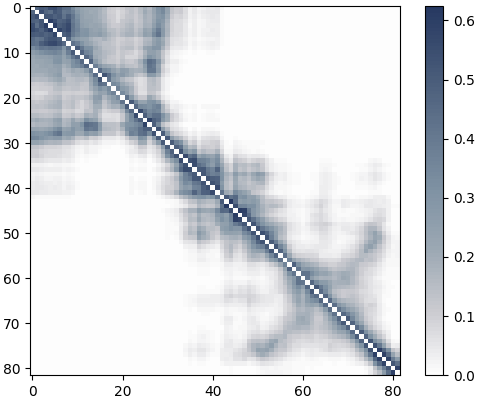}
        \caption{Replica Office 3}
    \end{subfigure}\\

    \begin{subfigure}{0.19 \linewidth}
        \centering 
        \includegraphics[width=\linewidth]{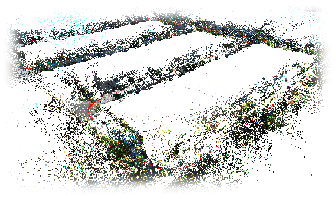}
        \caption{Outdoor Scene}
    \end{subfigure}
    \begin{subfigure}{0.19 \linewidth}
        \centering 
        \includegraphics[width=\linewidth]{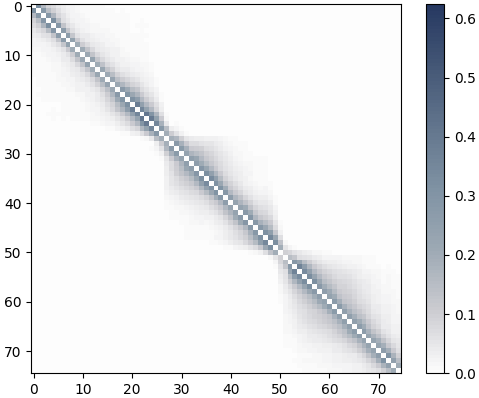}
        \caption{KITTI Seq. 00}
    \end{subfigure}
    \begin{subfigure}{0.19 \linewidth}
        \centering 
        \includegraphics[width=\linewidth]{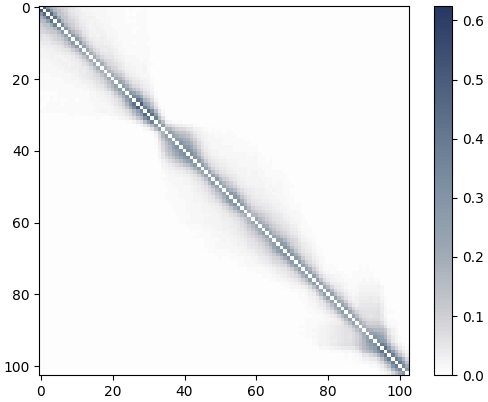}
        \caption{KITTI Seq. 05}
    \end{subfigure}
    \begin{subfigure}{0.19 \linewidth}
        \centering 
        \includegraphics[width=\linewidth]{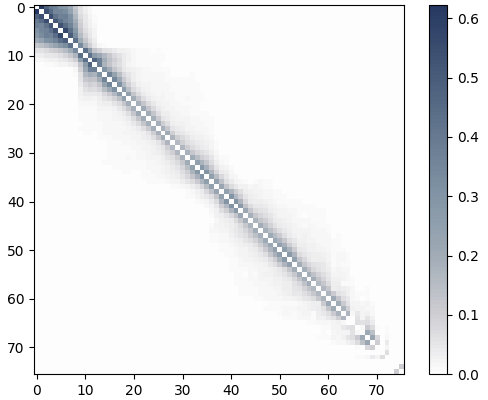}
        \caption{KITTI 360 Seq. 0000}
    \end{subfigure}
    \begin{subfigure}{0.19 \linewidth}
        \centering 
        \includegraphics[width=\linewidth]{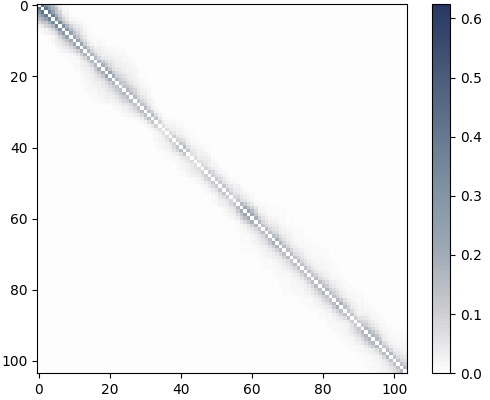}
        \caption{KITTI 360 Seq. 0004}
    \end{subfigure}

    \caption{The co-visibility matrix of DROID-SLAM for the KITTI, KITTI 360 and Replica dataset.}
    \label{fig:factor_graph}
\end{figure*}

\begin{figure*}[h]
    \centering
    \includegraphics[width=\linewidth]{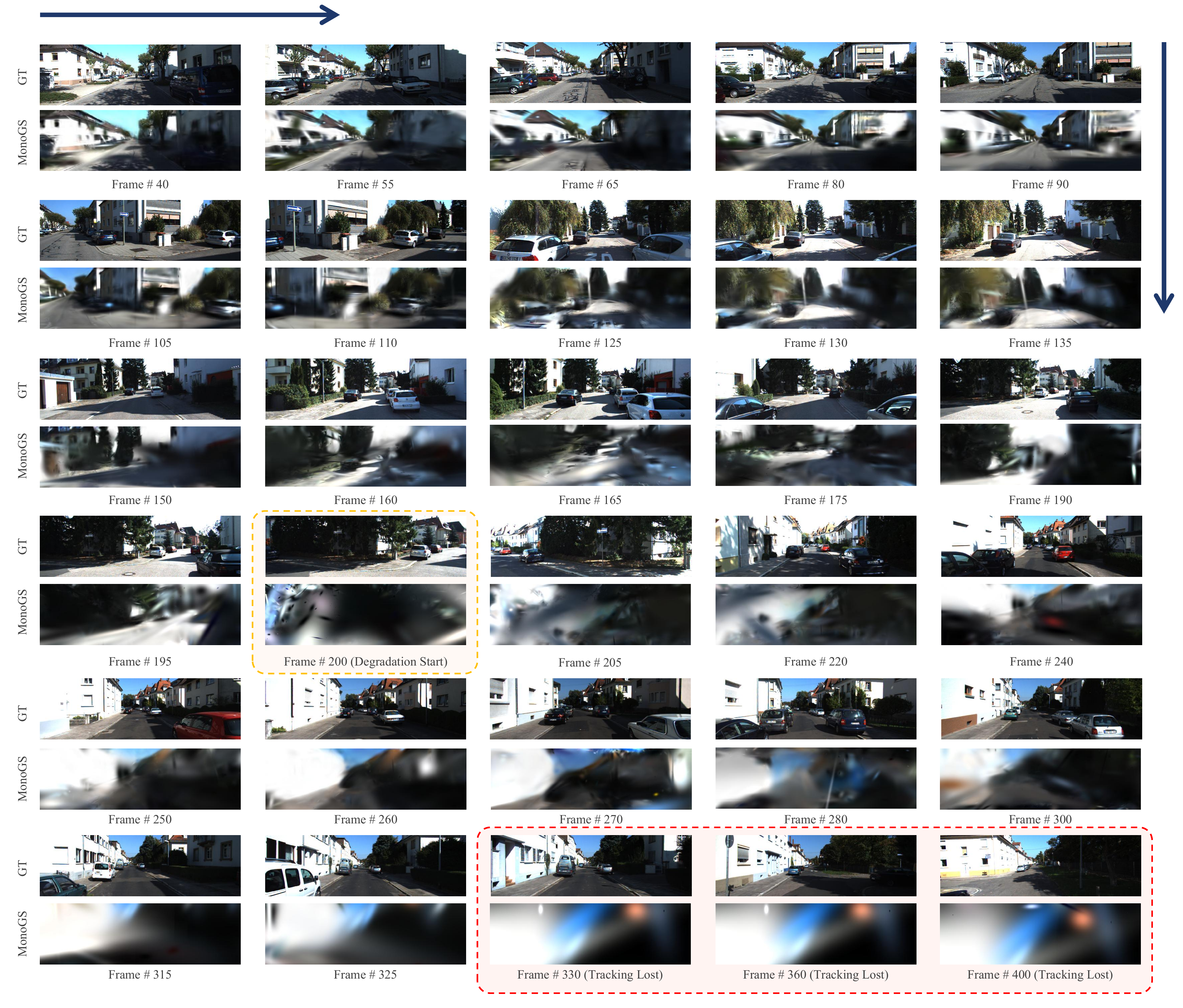}

    \caption{MonoGS on KITTI Seqence 00.}
    \label{fig:monogs-kitti-00}
\end{figure*}

Most monocular SLAM methods are well-validated on small-scale datasets but remain underexplored in large-scale outdoor long-sequence scenarios. DROID-SLAM \cite{droidslam}, though effective indoors, struggles with scale errors and computational overhead in outdoor datasets due to its reliance on dense bundle adjustment within a factor graph. MonoGS \cite{matsuki2024gaussian}, based on 3D Gaussian Splatting (3DGS) \cite{kerbl20233gs}, offers a promising alternative with its explicit representation, enabling high-fidelity mapping in unbounded environments. However, both approaches face significant challenges when applied to outdoor long-sequence datasets, as discussed in the following sections.

\subsection{DROID-SLAM based Methods}

DROID-SLAM \cite{droidslam}, introduced in 2021, represents a significant advancement in SLAM systems by making the entire SLAM pipeline fully differentiable. This innovation enabled the seamless integration of SLAM with deep learning techniques, outperforming traditional non-learning-based SLAM algorithms on smaller-scale datasets. Before DROID-SLAM, deep learning-based SLAM methods struggled to match the performance and reliability of classical approaches.

At its core, DROID-SLAM reformulates the SLAM problem as a joint optimization task, minimizing errors in pose estimation and map construction. It employs a recurrent iterative structure, leveraging the Turing-complete nature of recurrent neural networks (RNNs) \cite{siegelmann1991turing, siegelmann1992computational} to perform iterative optimization. This structure allows DROID-SLAM to iteratively refine dense correspondences and pose estimates, utilizing dense optical flow for robust matching between keyframes. Furthermore, DROID-SLAM integrates a dense bundle adjustment mechanism based on factor graph optimization, enabling accurate pose refinement without requiring explicit loop closure modules.

The system’s differentiable design and robust tracking capabilities have made it a foundational component in subsequent SLAM algorithms, such as GO-SLAM \cite{zhang2023go}, NeRF-SLAM \cite{rosinol2023nerf} and Splat-SLAM \cite{sandstrom2024splat}. These methods extend DROID-SLAM's principles, combining them with advanced mapping techniques like Neural Radiance Fields (NeRF) \cite{mildenhall2021nerf} or 3D Gaussian Splatting \cite{kerbl20233gs}. Such integrations have achieved state-of-the-art performance on indoor datasets, highlighting DROID-SLAM's adaptability and theoretical robustness.

However, DROID-SLAM encounters significant challenges in outdoor, long-sequence datasets like KITTI \cite{geiger2012kitti} and KITTI 360 \cite{liao2022kitti}, primarily due to its reliance on optical flow to construct the factor graph. The core of DROID-SLAM lies in performing Dense Bundle Adjustment (DBA) on this factor graph to optimize pose and map estimates. While effective in smaller-scale datasets with dense co-visibility, the factor graph’s sparsity in outdoor settings limits the effectiveness of this approach. Specifically, the co-visibility matrix in outdoor sequences has fewer edges due to the reliance on optical flow, which inherently reduces connections between frames. As a result, the optimization window for the DBA module in large-scale outdoor datasets is significantly smaller than in confined indoor scenarios.

This small optimization window amplifies the accumulation of scale drift, as errors that could otherwise be corrected through a larger window of jointly optimized frames remain unaddressed. Figure \ref{fig:factor_graph} illustrates this limitation by comparing the co-visibility graphs of DROID-SLAM on KITTI, KITTI 360 and Replica \cite{straub2019replica}. The co-visibility graph for KITTI reveals sparse connections, primarily linking each frame to its nearest 4 to 5 neighbors. In contrast, the graph for the indoor Replica dataset is significantly denser, reflecting DROID-SLAM’s inherent suitability for structured, small-scale environments.

In addition to the sparsity challenge, maintaining even a small factor graph and executing Dense Bundle Adjustment over kilometer-scale sequences comes at a steep computational cost. DROID-SLAM relies on implicit loop closure through inter-frame co-visibility relationships, which becomes less effective in outdoor scenarios where loops are infrequent and harder to detect. The computational resources required to process and optimize the factor graph in such scenarios are prohibitively high, further limiting its applicability to unbounded environments.


\subsection{MonoGS}

NeRF-based SLAM systems \cite{zhu2022nice, zhu2023nicer, zhang2023go} have shown great potential for high-quality scene reconstruction, particularly in indoor environments. However, they face significant limitations when applied to unbounded outdoor sequences. The primary challenge with NeRF lies in the slow training and rendering speeds, which make it difficult to process large-scale, long-sequence datasets in real-time. While recent techniques like Block-NeRF \cite{tancik2022block} and Mega-NeRF \cite{turki2022mega} have made strides in scalability, they are still not used for long-duration SLAM tasks, especially in outdoor environments.

In contrast, MonoGS \cite{matsuki2024gaussian} addresses one of these challenges by using 3D Gaussian Splatting (3DGS)  \cite{kerbl20233gs} instead of NeRF. The key advantage of 3DGS is its ability to represent scenes with smooth, continuously differentiable Gaussian blobs, which can be rendered efficiently at high frame rates. This allows for real-time mapping and tracking. By adopting 3DGS, MonoGS overcomes the slow training and rendering issues associated with NeRF-based methods, enabling high-fidelity, real-time SLAM performance with just monocular RGB input. MonoGS achieves this through innovations like an analytic Jacobian for pose optimization, isotropic shape regularization for geometric consistency, and a resource allocation strategy that maintains map accuracy without compromising efficiency.

Thus, MonoGS not only solves some of the key pain points of slow training and rendering inherent in NeRF but also provides a more scalable and efficient solution for  SLAM tasks. Due to its seamless integration of 3DGS into the SLAM framework, our method adopts MonoGS as its foundational codebase. MonoGS is particularly notable for its reliance on 3DGS as both the scene representation and the foundation of its tracking module. However, this heavy dependence on 3DGS rendering introduces significant challenges, particularly in outdoor long-sequence scenarios. The tracking process in MonoGS estimates camera poses by minimizing rendering losses, which inherently requires 3DGS to produce high-fidelity novel views. While this approach works well in indoor environments—where the relatively small scale, simple scene structures, and limited camera motion allow 3DGS to converge quickly—it becomes problematic in outdoor settings with larger scales and more complex geometries.

In outdoor scenarios, the convergence of 3DGS rendering is significantly slower due to the increased complexity and scale of the scenes. As a result, MonoGS often attempts to optimize pose tracking using a partially converged or inaccurate 3DGS map. This leads to pose estimation errors, which accumulate over time, especially during large-scale camera motions. Furthermore, in outdoor settings, the inaccurate depth of distant Gaussians introduces additional errors, making subsequent pose tracking increasingly unreliable. These issues create a feedback loop, where pose inaccuracies degrade the 3DGS map, which in turn further worsens pose estimation. Over long sequences, this cycle can eventually destabilize the entire system.

Figure \ref{fig:monogs-kitti-00} provides a clear illustration of these limitations. At the beginning of the sequence, although the mapping quality is suboptimal, the rendered scene remains recognizable, indicating that 3DGS is functioning adequately for the initial frames. However, as the sequence progresses, the cumulative pose errors and mapping inaccuracies cause the rendered scene to become increasingly distorted. By the second major turn, MonoGS’s tracking module is overwhelmed by these errors, leading to a complete breakdown of the system. The rendered scene at this stage is entirely unrecognizable, demonstrating the cascading failure of both the tracking and mapping components.

So \textbf{what challenges are we facing on outdoor long-sequence datasets?} Outdoor long-sequence datasets expose fundamental limitations in existing SLAM systems due to their expansive scale, complex scene geometries, and diverse camera motions. Systems like DROID-SLAM encounter scale drift and computational inefficiencies from sparse co-visibility and small optimization windows, while MonoGS struggles with slow 3D Gaussian Splatting convergence, leading to compounding pose and mapping inaccuracies

\section{Details about Pose Tracking}
\label{sec:rationale}
To estimate camera motion between images, we begin by extracting image features using the DISK network \cite{tyszkiewicz2020disk}, which provides robust descriptors capable of capturing rich feature representations. These features are then matched using LightGlue \cite{lindenberger2023lightglue}, a state-of-the-art deep learning-based matcher. By leveraging adaptive filtering and dynamic weighting strategies, LightGlue establishes reliable correspondences, even in challenging scenarios with low texture or significant viewpoint changes.

We adopt the methodology from DF-VO \cite{zhan2021df} for estimating the poses with fundamental matrix $\mathbf{F}$ or essential matrix $\mathbf{E}$ \cite{bian2019evaluation, nister2004efficient, zhang1998determining}. With the matched feature points $\mathbf{p}_i$ and $\mathbf{p}_j$, these matrices 
 could be computed using the classical epipolar constraint:

\begin{equation}
\mathbf{F} = \mathbf{K}^{-T} \mathbf{E} \mathbf{K}^{-1}, \quad \mathbf{E} = [t] \times \mathbf{R}.
\end{equation}

Here, $\mathbf{p}_i$ and $\mathbf{p}_j$ represent the homogeneous coordinates of corresponding points in the two images, expressed as $\mathbf{p} = [u, v, 1]^T$, where $(u, v)$ are the pixel coordinates. The epipolar constraint is enforced as:

\begin{equation}
\mathbf{p}_j^T \mathbf{K}^{-T} \mathbf{E} \mathbf{K}^{-1} \mathbf{p}_i = 0.
\end{equation}

The camera motion $[\mathbf{R}, \mathbf{t}]$ is recovered by decomposing $\mathbf{F}$ or $\mathbf{E}$.

While effective in many scenarios, this approach can encounter challenges under certain conditions, such as motion degeneracy (e.g., pure rotation) or scale ambiguity inherent in the essential matrix.

To refine camera pose estimation, the Perspective-n-Point (PnP) algorithm is employed \cite{zhan2021df}, which minimizes the reprojection error using 3D-2D correspondences:

\begin{equation}
e = \sum \Vert \mathbf{K}(\mathbf{R}\mathbf{X}_i + \mathbf{t}) - \mathbf{p}_j \Vert^2.
\end{equation}

The required 3D information for 3D-2D correspondences is derived from dense depth maps extracted using the UniDepth model \cite{piccinelli2024unidepth}. By providing depth estimates from monocular RGB images, UniDepth ensures a reliable representation of the scene’s structure, mitigating depth ambiguity in monocular setups.

To enhance robustness, we adopt the geometric robust information criterion (GRIC) \cite{zhan2021df, torr1999problem} as a model selection strategy. GRIC evaluates the suitability of essential matrix decomposition, identifying cases of motion or structure degeneracy. The GRIC function is defined as:

\begin{equation}
\text{GRIC} = \sum \rho(e_i^2) + \log(4)dn + \log(4n)k,
\end{equation}

with

\begin{equation}
\rho(e_i^2) = \min\left(\frac{e_i^2}{2(r - d)\sigma^2}, 1\right).
\end{equation}

Here, $d$ is the structure dimension, $n$ is the number of matched features, $k$ is the number of motion model parameters, and $\sigma$ is the standard deviation of measurement error. When GRIC$_F$ exceeds GRIC$_H$, we switch to PnP, utilizing UniDepth-derived depth information for improved pose estimation.

By combining robust feature matching with LightGlue, dense depth estimation from UniDepth, and GRIC-based model selection, our system addresses the limitations of traditional epipolar geometry pipelines, achieving improved resilience and accuracy in monocular setups.

\section{Loop Correction}

In our system, we integrate proximity-based loop closure detection with a traditional SLAM back-end, using image retrieval techniques to identify and correct loop closures, particularly for enhancing long-term localization accuracy. For this, we utilize DBoW2 \cite{galvez2012bags} for image retrieval by detecting candidate image pairs that suggest a loop closure, extracting ORB \cite{rublee2011orb} features from each frame. These feature extraction, indexing, and search operations occur concurrently in a separate thread, minimizing runtime overhead. Additionally, Non-Maximum Suppression (NMS) to prevent overly frequent detections referred to \cite{dpv-slam}. We perform a Sim(3) optimization for global pose estimates by optimizing a smoothness term and loop closure constraints using the Levenberg-Marquardt algorithm. The loop correction method closely follows the work of \cite{dpv-slam, strasdat2010scale}. This method is a classic Sim(3)-based optimization approach that has been applied to various SLAM methods over the past fifteen years. Given the $\text{SE}(3)$ poses of all keyframes, suppose a loop closure is detected between frame $j$ and frame $k$. Define the similarity transformation as $S_i=(t_i,R_i,s_i) \in \text{Sim}(3)$, and compute the residual between the two frames as:
\begin{equation}
r_{jk}=\log_{\text{Sim}(3)}(\Delta S_{jk}^{\text{loop}}\cdot S_{j}^{-1}\cdot S_{k}).
\end{equation}
Without an explicit pose factor graph, a virtual factor graph can be constructed by considering only the connections between adjacent frames. The residual between consecutive frames is defined as:
\begin{equation}
r_{i}=\log_{\text{Sim}(3)}(\Delta S_{i,i+1}^{-1}\cdot S_{i}^{-1}\cdot S_{i+1}).
\end{equation}
The objective function for optimization is then formulated as:
\begin{equation}
\underset{S_1,\cdots,S_N}{\arg \min} \sum_{i}^{N} \Vert r_i\Vert^2_2 + \sum_{(j,k)}^L \Vert r_{jk}\Vert^2_2.
\end{equation}
Expanding the residual terms gives:
\begin{equation}
\begin{aligned}
\underset{S_1,\cdots,S_N}{\arg \min} \sum_{i}^{N} \Vert \log_{\text{Sim}(3)}(\Delta S_{i,i+1}^{-1}\cdot S_{i}^{-1}\cdot S_{i+1})\Vert^2_2 \\
+ \sum_{(j,k)}^L \Vert \log_{\text{Sim}(3)}(\Delta S_{jk}^{\text{loop}}\cdot S_{j}^{-1}\cdot S_{k})\Vert^2_2.
\end{aligned}
\end{equation}

This objective is optimized using the Levenberg-Marquardt (LM) algorithm. After optimization, the updated similarity transformations $S_i$ are used to update the global poses as $G_i \leftarrow (t_i, R_i)$.

To simplify the optimization, the objective function reduces to:
\begin{equation}
\min \Vert r(S) \Vert^2 \rightarrow \min \Vert\log_{\text{Sim}(3)}(\Delta S_{jk}^{\text{loop}}\cdot S_{j}^{-1}\cdot S_{k})\Vert^2.
\end{equation}
At each optimization step, the relative transformation $\Delta S_{jk}^{\text{loop}}$ is first computed and treated as a constant $C$:
\begin{equation}
r= \Vert\log_{\text{Sim}(3)}(C\cdot S_{j}^{-1}\cdot S_{k})\Vert^2.
\end{equation}
The Jacobian matrix is then computed as:
\begin{equation}
J_j = \frac{\partial r}{\partial S_j},\quad J_k = \frac{\partial r}{\partial S_k},\quad J=[J_j,\, J_k].
\end{equation}
The update increment $\Delta S=[\Delta S_j, \Delta S_k]$ is estimated using a first-order Taylor approximation:
\begin{equation}
r \approx r + J\Delta S.
\end{equation}
Thus, the optimization problem reduces to:
\begin{equation}
\begin{aligned}
\Delta S &= \arg \min \Vert r + J\Delta S\Vert^2 \\
&= \arg \min ( r + J\Delta S)^\top( r + J\Delta S).
\end{aligned}
\end{equation}
Expanding the expression:
\begin{equation}
\Delta S = \arg \min ( \Vert r \Vert^2 + \Delta S^\top J^\top J \Delta S + 2\Delta S^\top J^\top r).
\end{equation}
Taking the derivative with respect to $\Delta S$ and setting it to zero leads to:
\begin{equation}
J^\top J\Delta S = -J^\top r.
\end{equation}
To prevent divergence due to large steps, a damping term $\lambda\, \text{diag}(J^\top J)$ is added, along with a small regularization term $\epsilon I$ to ensure numerical stability:
\begin{equation}
(J^\top J+\lambda\, \text{diag}(J^\top J)+\epsilon I)\Delta S = -J^\top r.
\end{equation}
This results in solving a linear system of the form:
\begin{equation}
A\Delta x=b.
\end{equation}

In the next iteration, the updated $\text{Sim}(3)$ transformations are used to recompute $\Delta S_{jk}^{\text{loop}}$, and the process continues iteratively until convergence.

\begin{figure*}[t]
  \centering
   \includegraphics[width=0.75\linewidth]{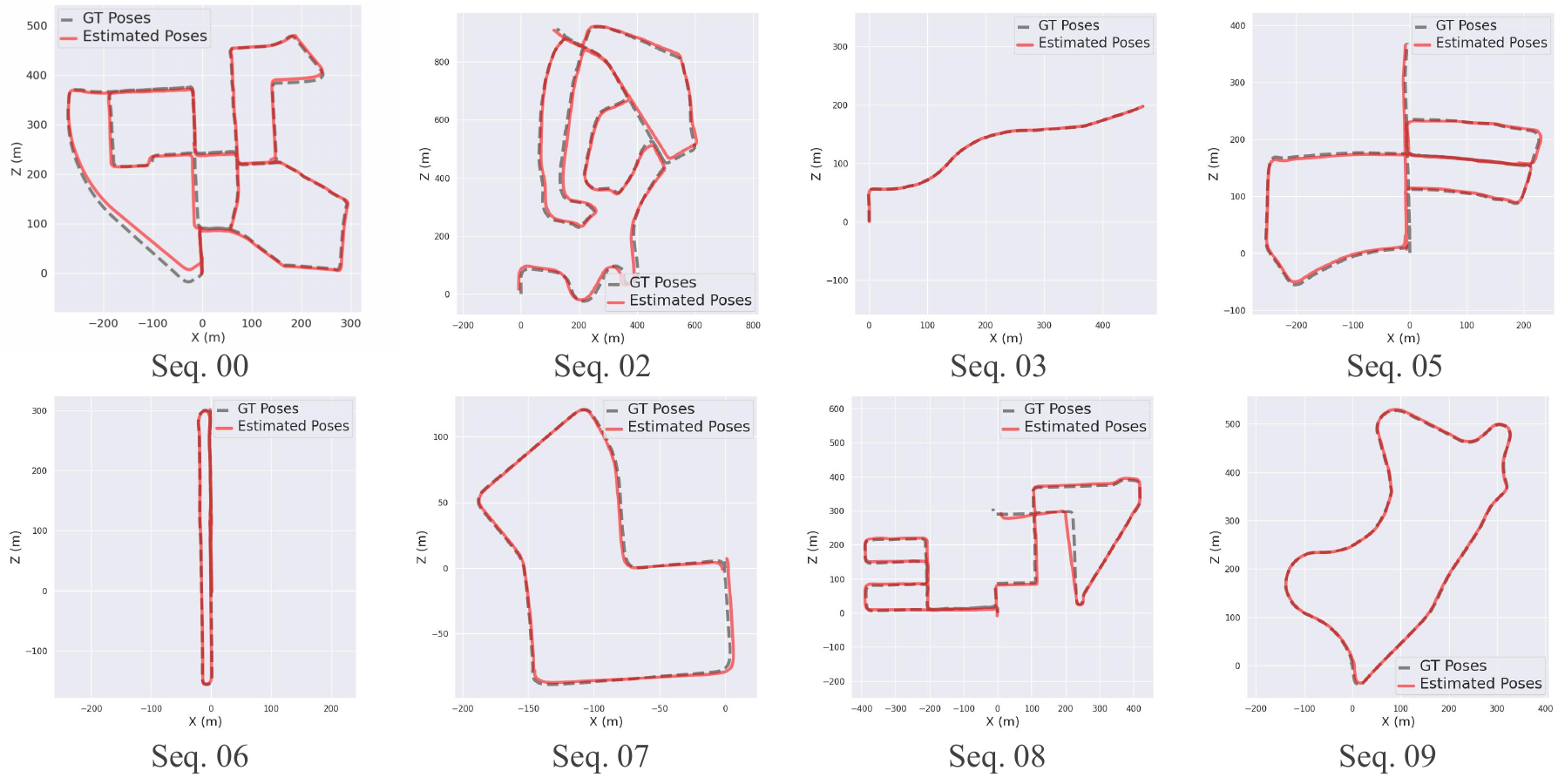}
   \caption{Camera trajectory visualization for the KITTI dataset.}
   \label{fig:kitti_all_traj}
\end{figure*}

\begin{figure*}[t]
  \centering
   \includegraphics[width=1\linewidth]{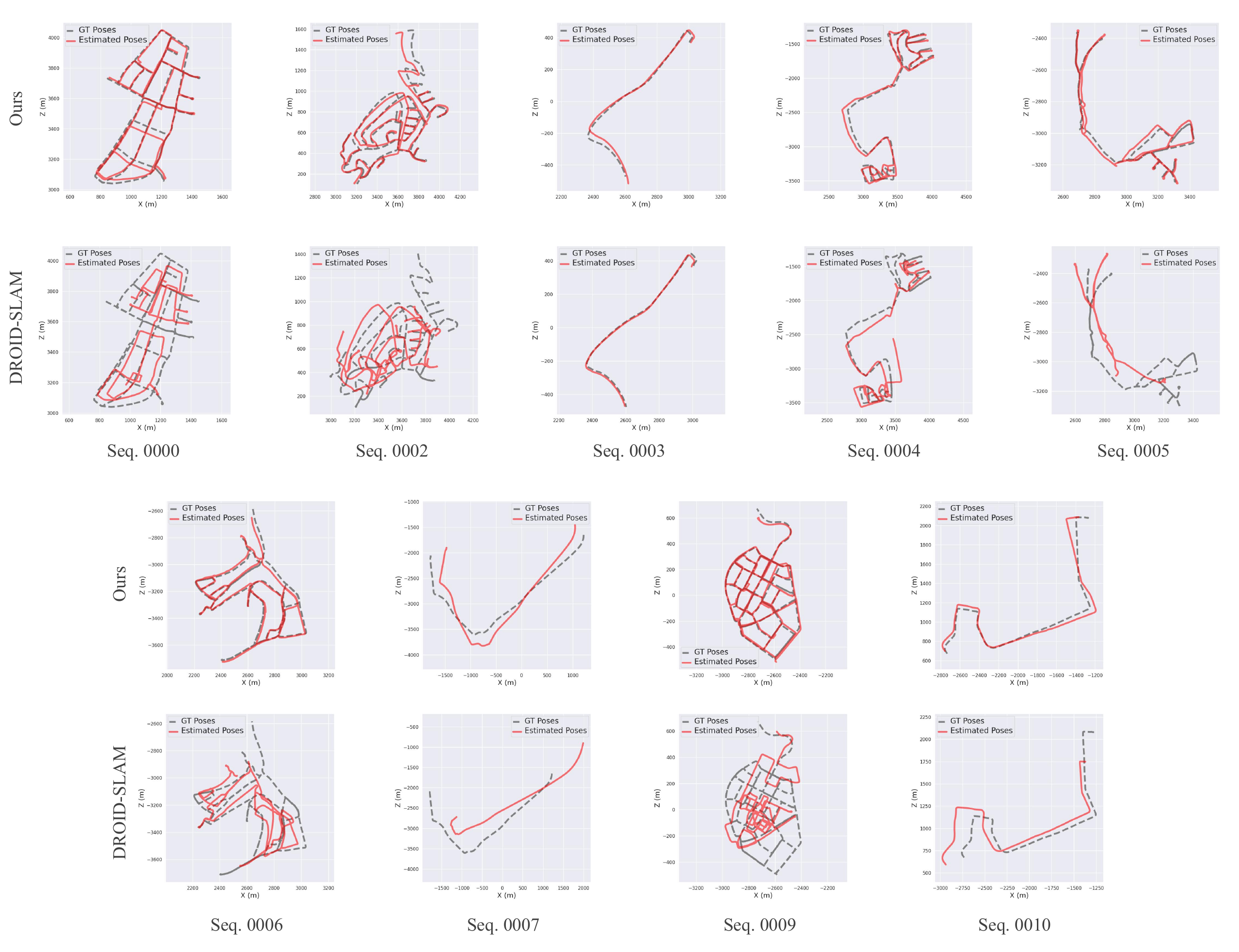}
   \caption{Camera trajectory visualization for the KITTI 360 dataset.}
   \label{fig:kitti_360_all_traj}
\end{figure*}

\begin{figure*}[htbp]
    \centering
    \includegraphics[width=\linewidth]{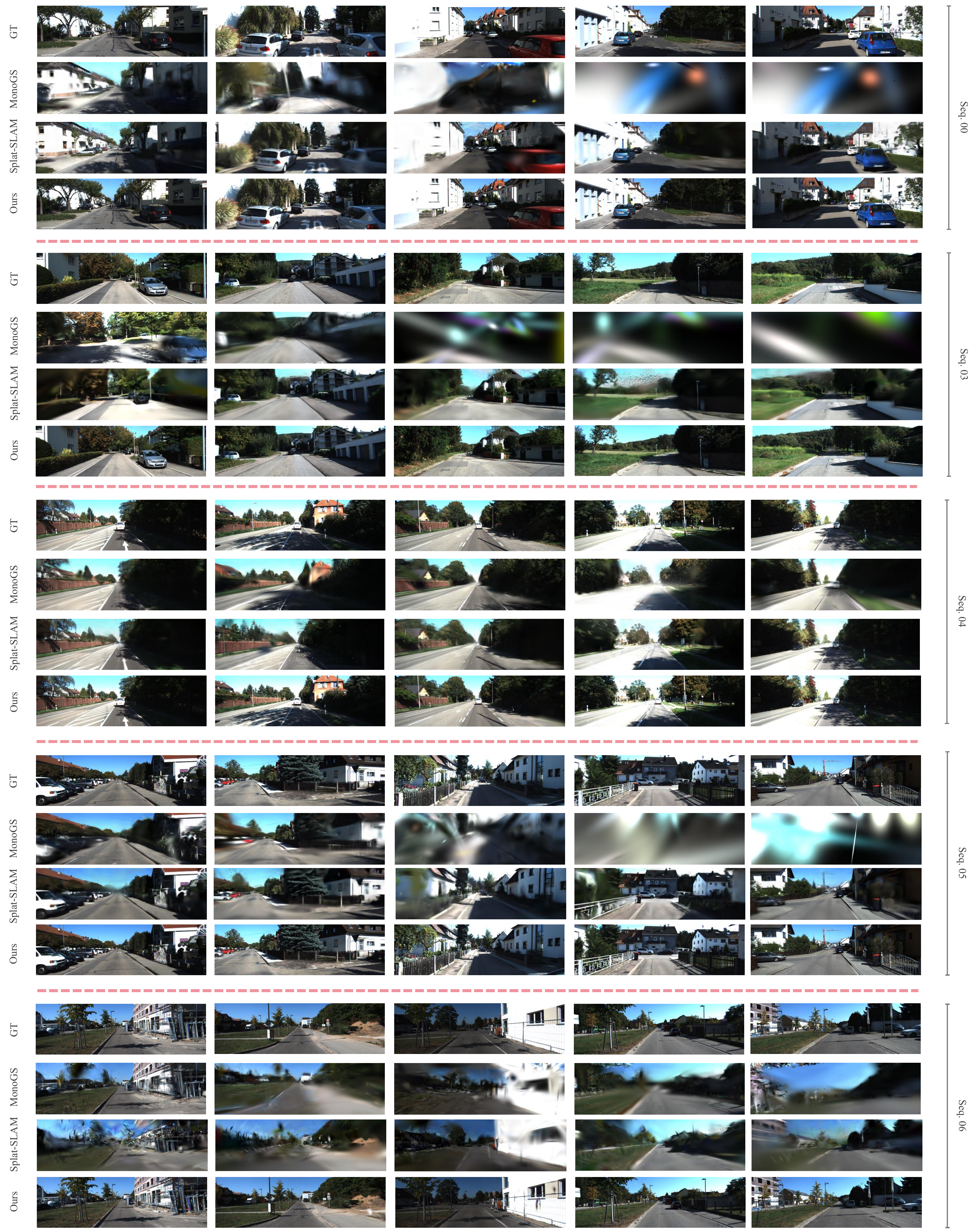}
    \caption{Rendering result for KITTI dataset of Seq. 00 to 06.}
    \label{fig:kitti_full_render1}
\end{figure*}

\begin{figure*}[htbp]
    \centering
    \includegraphics[width=\linewidth]{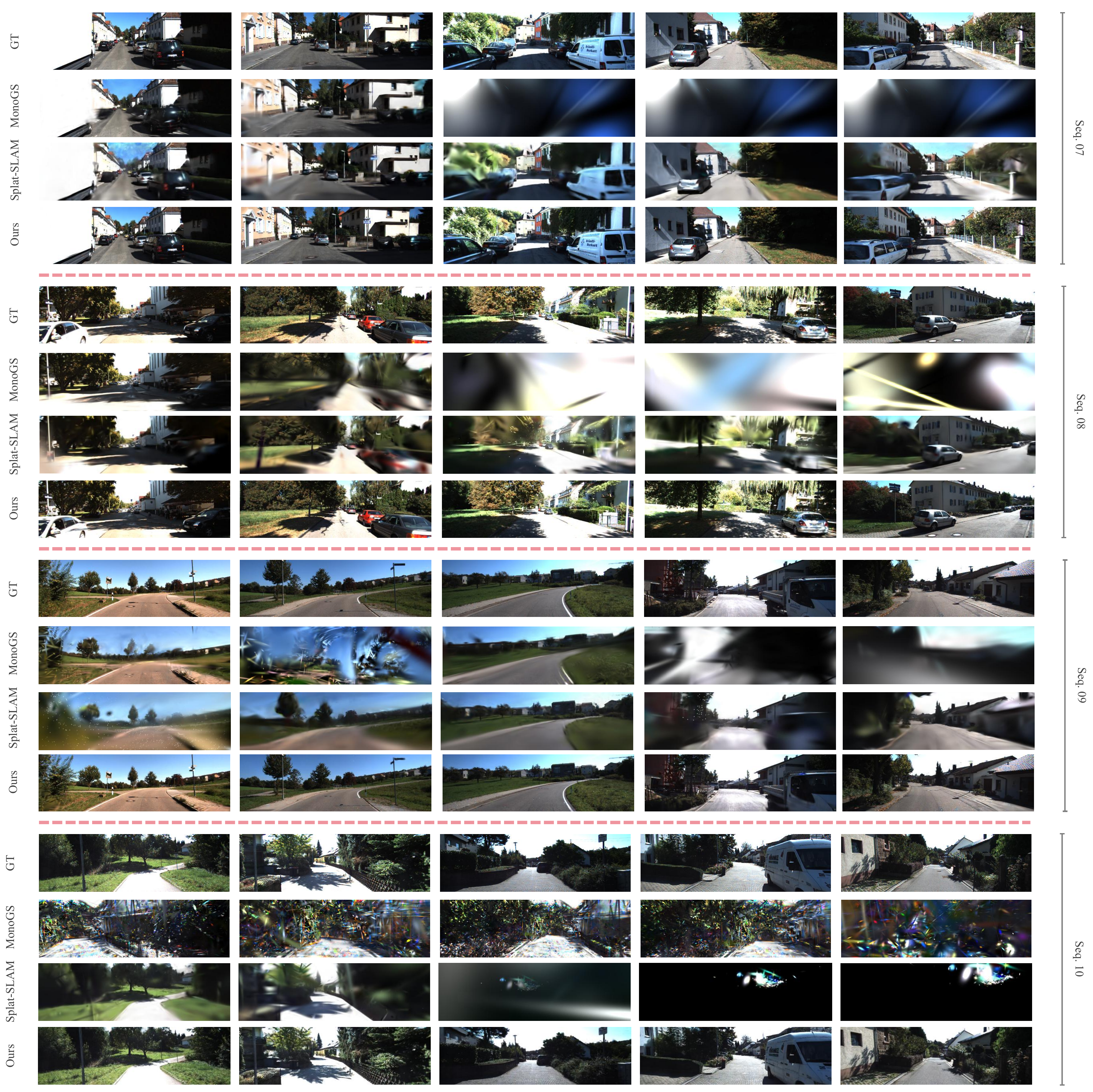}
    \caption{Rendering result for KITTI dataset of Seq. 07 to 10.}
    \label{fig:kitti_full_render2}
\end{figure*}

\section{Visualization of Camera Trajectory on KITTI and KITTI 360 Dataset.}

The main paper presents trajectory comparisons for the KITTI \cite{geiger2012kitti} and KITTI 360 \cite{liao2022kitti} datasets, respectively. This appendix provides a detailed discussion of these visualizations, further illustrating the performance of our method in maintaining accurate and stable camera pose estimates across long sequences.

For KITTI, our method demonstrates consistent trajectory alignment with ground truth across challenging sections of the sequence as demonstrated in Figure \ref{fig:kitti_all_traj}. For KITTI 360, Figure \ref{fig:kitti_360_all_traj} provides a more comprehensive evaluation of ultra-long trajectories, spanning up to 14,607 frames. Notably, DROID-SLAM \cite{droidslam} achieves competitive performance in sequence 0003, where its trajectory slightly outperforms ours. However, across the majority of sequences, DROID-SLAM exhibits substantial scale drift, consistent with the challenges described in Section \ref{sec:challenges}. These large-scale deviations undermine its ability to provide reliable pose estimates over extended sequences. In contrast, our method maintains stable and accurate camera poses throughout, highlighting its resilience to error accumulation and scalability to unbounded scenarios. We also present the rendered images in Figure \ref{fig:kitti_full_render1} and \ref{fig:kitti_full_render2} of our approach alongside those of MonoGS \cite{matsuki2024gaussian} on the KITTI dataset. Consistent with the discussion in Section \ref{sec:challenges}, MonoGS performs poorly on long-sequence outdoor datasets such as KITTI, whereas our method demonstrates robust performance.

Overall, these visualizations underscore the adaptability and robustness of our system across both traditional and ultra-long outdoor SLAM tasks, with detailed trajectory plots revealing its superior performance in maintaining trajectory fidelity.

Our paper closely follows a series of recent works, such as \cite{droidslam, keetha2024splatam, zhang2023go, zhu2022nice, zhu2023nicer, dpv-slam, dpvo, matsuki2024gaussian, sandstrom2024splat, sandstrom2023point} et al., which use ATE as a metric for evaluating tracking accuracy. However, we also note that some more earlier works \cite{zhan2021df} use other metrics, such as translation and rotation drift over segment lengths of 100 to 800 meters with loop closure disabled. Since recent works have not adopted this metric and ATE is a better measure of long-term tracking performance in long outdoor sequences, we report ATE in the main paper and provide T/R Drift data in the Table \ref{tab:t/r_drift} for reference.

\quad

\begin{table}[h]
\centering
\resizebox{0.95\linewidth}{!}{
\begin{tabular}{c|c|ccccccccccc}
\toprule
\textbf{Methods} & \textbf{Metric} & \textbf{00} & \textbf{01} & \textbf{02} & \textbf{03} & \textbf{04} & \textbf{05} & \textbf{06} & \textbf{07} & \textbf{08} & \textbf{09} & \textbf{10} \\
\midrule
ORB SLAM w/o LC       & $t_{err}$               & 11.43       & 107.57      & 10.34       & 0.97        & 1.3         & 9.04        & 14.56       & 9.77        & 11.46       & 9.3         & 2.57        \\
                     \cite{orbslam}& $r_{err}$               & 0.58        & 0.89        & 0.26        & 0.19        & 0.27        & 0.26        & 0.26        & 0.36        & 0.28        & 0.26        & 0.32        \\
                     \midrule
DF-VO                & $t_{err}$               & 2.33        & 39.46       & 3.24        & 2.21        & 1.43        & 1.09        & 1.15        & 0.63        & 2.18        & 2.4         & 1.82        \\
                     \cite{zhan2021df}& $r_{err}$               & 0.63        & 0.5         & 0.49        & 0.38        & 0.3         & 0.25        & 0.39        & 0.29        & 0.32        & 0.24        & 0.38        \\
                     \midrule
MonoGS               & $t_{err}$               & /           & 99.4        & /           & /           & 7.34        & /           & 101.73      & /           & /           & /           & /           \\
                     \cite{matsuki2024gaussian}& $r_{err}$               & /           & 27.02       & /           & /           & 3.57        & /           & 10.82       & /           & /           & /           & /           \\
                     \midrule
Splat-SLAM           & $t_{err}$               & 17.97       & /           & /           & 2.41        & 33.78       & 5.4         & 33.28       & 10.05       & 12.67       & 7.13        & 30.82       \\
                     \cite{sandstrom2024splat}& $r_{err}$               & 1.62        & /           & /           & 0.33        & 0.56        & 0.48        & 0.49        & 0.42        & 0.72        & 0.24        & 3.71        \\
                     \midrule
Ours w/o LC                 & $t_{err}$               & 1.38        & 41.08       & 1.49        & 2.21        & 2.23        & 1.49        & 1.6         & 1.06        & 2.29        & 1.05        & 1.19        \\
                     & $r_{err}$               & 0.43        & 0.92        & 0.39        & 0.58        & 0.19        & 0.39        & 0.39        & 0.47        & 0.61        & 0.28        & 2.27   \\
\bottomrule
\end{tabular}
}
\caption{Translation and rotation drift of different methods}
\label{tab:t/r_drift}
\end{table}

\section{Limitation of our work}

Currently, research on kilometer-scale outdoor monocular RGB SLAM using NeRF \cite{mildenhall2021nerf} or 3DGS \cite{kerbl20233gs} is still in its nascent stages. Our method is specifically tailored for autonomous driving scenarios and places less emphasis on other scene types. Our approach is not the best solution for indoor environments. Moreover, limitations such as motion blur, camera shake, glare, overexposure, and low-texture scenes can reduce tracking accuracy, though these challenges are more pronounced in non-driving scenarios.  Additionally, the memory requirements of NeRF or 3DGS present challenges for city-scale scenes.

Future work could explore solutions to these issues, extending applicability beyond driving-focused datasets and further improving robustness in various types of environments.

\end{document}